\documentclass[acmlarge,screen]{acmart}

\usepackage{graphicx}
\usepackage{subfigure}
\graphicspath{{pictures/}}
\usepackage{booktabs}    
\usepackage{multirow}    
\usepackage{tabularx}

\usepackage{amssymb}             
\usepackage{amsfonts}            
\usepackage{mathrsfs}            
\usepackage{array}
\usepackage{algorithm}
\usepackage{algorithmic}

\def\BibTeX{{\rm B\kern-.05em{\sc i\kern-.025em b}\kern-.08emT\kern-.1667em\lower.7ex\hbox{E}\kern-.125emX}}
    
\copyrightyear{2020}
\acmYear{2020}
\setcopyright{acmlicensed}
\acmConference[Woodstock '18]{Woodstock '18: ACM Symposium on Neural Gaze Detection}{June 03--05, 2018}{Woodstock, NY}
\acmBooktitle{Woodstock '18: ACM Symposium on Neural Gaze Detection, June 03--05, 2018, Woodstock, NY}
\acmPrice{15.00}
\acmDOI{10.1145/1122445.1122456}
\acmISBN{978-1-4503-9999-9/18/06}

\begin{document}

%
\title{DeepExpress: Heterogeneous and Coupled Sequence Modeling \\ for Express Delivery Prediction}

%
\author{Siyuan~Ren}
\email{rensiyuan@mail.nwpu.edu.cn}
\affiliation{%
  \institution{Northwestern Polytechnical University}
}

\author{Bin~Guo}
\authornotemark[1]
\affiliation{%
  \institution{Northwestern Polytechnical University}
}
\email{guobin.keio@gmail.com(Corresponding author)}

\author{Longbing~Cao}
\email{longbing.cao@uts.edu.au}
\affiliation{%
  \institution{University of Technology Sydney}
}

\author{Ke~Li}
\email{kei1992@mail.nwpu.edu.cn}
\affiliation{%
  \institution{Northwestern Polytechnical University}
}

\author{Jiaqi~Liu}
\email{jqliu@nwpu.edu.cn}
\affiliation{%
  \institution{Northwestern Polytechnical University}
}

\author{Zhiwen~Yu}
\email{zhiwenyu@nwpu.edu.cn}
\affiliation{%
  \institution{Northwestern Polytechnical University}
}

%
\renewcommand{\shortauthors}{S.Ren et al.}

%
\begin{abstract}
The prediction of express delivery sequence, i.e., modeling and estimating the volumes of daily incoming and outgoing parcels for delivery, is critical for online business, logistics, and positive customer experience, and specifically for resource allocation optimization and promotional activity arrangement. A precise estimate of consumer delivery requests has to involve sequential factors such as shopping behaviors, weather conditions, events, business campaigns, and their couplings. Besides, conventional sequence prediction assumes a stable sequence evolution, failing to address complex nonlinear sequences and various feature effects in the above multi-source data. Although deep networks and attention mechanisms demonstrate the potential of complex sequence modeling, extant networks ignore the heterogeneous and coupling situation between features and sequences, resulting in weak prediction accuracy. To address these issues, we propose DeepExpress - a deep-learning based express delivery sequence prediction model, which extends the classic seq2seq framework to learning complex coupling between sequence and features. DeepExpress leverages an express delivery seq2seq learning, a carefully-designed heterogeneous feature representation, and a novel joint training attention mechanism to adaptively map heterogeneous data, and capture sequence-feature coupling for precise estimation. Experimental results on real-world data demonstrate that the proposed method outperforms both shallow and deep baseline models.
\end{abstract}

%
%


%

%

%
\maketitle

\section{Introduction}
\label{sec:IN}
With the rapid growth of e-commerce, the way of online shopping has not only brought about commercial changes but also changed the definitions of consumers and their shopping behaviors, logistic requests, and relevant businesses. To leverage online businesses and activities, express delivery service plays an increasingly important role in one's digitalized life and bridges the gap between virtual online commerce spaces and real societies. Indeed, express delivery service reshapes both commerce and people's behaviors.

\begin{figure}[!t]
	\centering
	\includegraphics[width=0.75\linewidth]{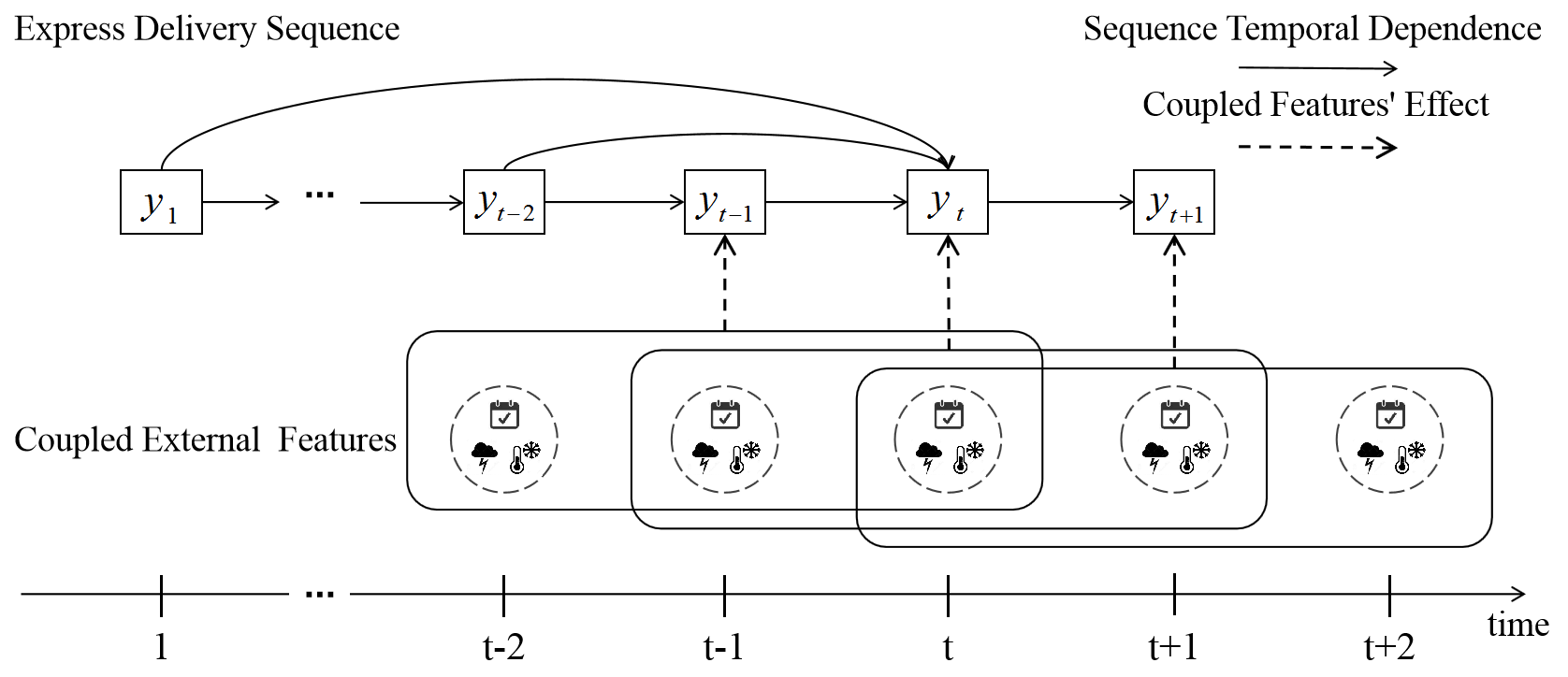}
	\caption{A perspective of express delivery prediction: it is a complicated problem that involves sequence temporal dependence and external features' effect. Specifically, modeling features' effect is challenging as the heterogeneous and complex coupling between features and sequences}
	\label{fig:1}
\end{figure}

A critical demand is to express delivery sequence prediction, i.e., estimating the number of incoming and outgoing parcels in an urban area on the historical multi-sourced urban data. This is challenging as it has to jointly model multiple historical and present factors that are coupled with each other, as shown in Fig. \ref{fig:1}, e.g., the delivery demand, consumer behaviors, resource allocation, and business activities. For example, a total of 272.5 million parcels were delivered in Xi'an (a city in China) from Jan. to Sept. 2018 with an average daily volume of 99 thousand\footnote{http://s.askci.com/data/area/610000/}. Bad planning of delivering such a large number of parcels would incur problems like delivery delay, resource waste, or overstock. In contrast, a precise prediction of delivery volume can inform effective business planning and activities, positive shopping experience, and efficient delivery strategies. 

Conventional approaches for express delivery sequence prediction are to establish a mathematical model to fit the historical time trend curve such as Autoregressive Integrated Moving Average (ARIMA) \cite{box1970distribution}, Bayesian Ridge Regression \cite{shi2016bayesian}, Support Vector Regression (SVR) \cite{bao2014multi}, and Prophet \footnote{https://facebook.github.io/prophet/}. However, such mathematical models rely on a stable trend and periodic change, and perform poorly with complicated nonlinear relations. Moreover, as shown in Fig. \ref{fig:1}, the delivery sequence changes as a joint effect of sequence auto-regressive and coupled external features' effect. The existing approaches often ignore such multiple aspects of coupling relations \cite{cao2015coupling}, which show important yet difficult to model for problems with multiple coupled behavior sequences \cite{cao2011coupled} or factors \cite{tkde_ZhuCLYK18} by either shallow or deep models \cite{ijcai_ZhangCZLS18, JianPCLG19}.

To model sequences with hidden relations, deep learning approaches such as Recurrent Neural Networks (RNNs) \cite{elman1991distributed}, Long Short-Term Memory units (LSTM) \cite{gers1999learning}, and the Gated Recurrent Unit (GRU) \cite{chung2014empirical} achieve great success. New research incorporates more advanced networks such as encoder-decoder architecture \cite{sutskever2014sequence} and attention mechanism \cite{vaswani2017attention} for time series prediction. However, existing approaches are still inefficient and inaccurate in practice due to the two following drawbacks:

a) Failing to take advantage of heterogeneous data. Existing approaches (e.g. CoST-Net \cite{ye2019co}) weaken the heterogeneous situations in the data, e.g., different data types, distributions, ranges, couplings \cite{cao2008domain}, rather than simply treating features equally and directly embedding them into a network. Therefore, much vital information will be lost in the process, and even worse, wrong features captured will cause performance degradation.   

b) Failing to model complex coupling relationship. Many researches \cite{liang2018geoman, ruan2020dynamic, zhang2020semi} considers the effects of external factors in the sequence prediction problem. However, it directly fuses them as discrete features in the forecasting phase, and omits their coupling relations. The express delivery sequence changes as a joint effect of sequence auto-regressive and coupled features' effect. The definition of coupling refers to the extensive associations between data, which contains a variety of relationships. In this paper, we explore two types of the coupling relationships between sequence and features that rare research quantifies before. Specifically, the coupling relationships mentioned in this paper mainly includes two aspects: 
\vspace{-0.3em}
\begin{itemize}
    \item{The sequence is affected by features not only in the current time but also the cumulative change over a period of time. For example, snow, holiday, and promotion events in the past or next few days will cause an unusual change in the delivery sequence.}
    \item{The correlation between each feature and sequence changes dynamically over time. It is critical to understand which features at which time has a greater effect on the prediction results, and this is helpful to improve model performance and interpretability.}
\end{itemize}
\vspace{-0.3em}

\setlength{\parindent}{1em}It is still a great challenge for deep models to address all these issues together. Thence this paper proposes DeepExpress, a novel deep-learning based express delivery sequence prediction model, which extend the classic seq2seq framework to learning complex coupling between sequence and features. DeepExpress leverages an express delivery seq2seq learning, a carefully-designed heterogeneous feature representation, and a novel joint training attention mechanism to adaptively map heterogeneous data, and capture sequence-feature coupling for precise estimation.

Accordingly, the major contributions of this work include:

1) We design a novel heterogeneous feature representation to adaptively map heterogeneous features to a unified hidden representation. Different from existing methods, heterogeneous features are treated differently through designed algorithms to preserve original information

2) An express delivery sequence prediction model DeepExpress is built on a joint training attention mechanism, which differs from general attention-based models to train different attentions collaboratively to learn the complex coupling between sequence and features.

3) We conduct extensive experiments on real-world express delivery data and compare our method with both shallow and deep learning based state-of-the-art methods. The evaluation results demonstrate that DeepExpress outperforms all the comparison baselines in different kinds of application scenarios.

The rest of the paper is organized as follows. In Section \ref{sec:RW}, we first ground the proposed DeepExpress model on formally reviewing the related work. Section \ref{sec:PD} formulates the problem and explains the challenges of heterogeneity and coupling in this paper. Section \ref{sec:MD} presents the framework of DeepExpress and delves into the core idea of the proposed model by formally denoting its key operations. Section \ref{sec:EV} evaluates results on real-world express delivery data and compares the performance of the proposed model with different baselines. Finally, section \ref{sec:CON} concludes the paper and offers promising future directions for research.

\section{Related Work}
\label{sec:RW}
Express delivery sequence prediction can be viewed as one of the applications of time series prediction problems \cite{mahalakshmi2016survey}. Therefore, we summarize the related work from two aspects, i.e., express delivery sequence prediction and time series prediction methods. Typically, we review many time series prediction methods, which are not limited to express delivery sequence prediction. 

\subsection{Express Delivery Sequence Prediction}
There are some previous research \cite{hou2005method, yin2016data, xie2014combined, sun2011forecast, tang2015based} on predicting express delivery sequences. Pang et al. \cite{pang2007forecasting} established a nonlinear prediction model of SVM for regional logistics demand prediction. Ming et al. \cite{ming2010forecasting} proposed a wavelet-based neural network to reveal the inherent non-linear mapping relationship between the regional economy and regional logistics demand. These studies distinguish different administrative divisions of a city. However, the region they considered is coarse-grained. Hu et al. \cite{hu2016method} constructed a logistics demand forecasting model by using economic indicators. Although there is a certain correlation between express delivery and economy, it only reflects the macro development trend of express delivery but does not directly reveal the changing rules, which leads to limited accuracy and reliability. Besides, Xie et al. \cite{xie2014combined} combined trend extrapolation, the grey model with the regression method by a genetic algorithm to determine the weight for each single prediction method. Zhou et al. \cite{zhou2012adaptive} used the optimal weighting combination algorithm for logistics volume prediction, which combines multiple linear regression, exponent smoothing, and the grey model and calculates the optimal weighting coefficients for each method. In summary, most of the above work addresses the macroscopic and long-term prediction (at the quarterly or yearly level). Ours is the first work that employs the seq2seq-based deep learning approach for practical express delivery estimation. Specifically, we focus on the fine-grained prediction, that is, forecasting the daily express volume in the target urban areas, which is different from their works in problem setting.

\subsection{Time Series Prediction Methods}
\subsubsection{Conventional methods}
Exponential smoothing \cite{gardner1985exponential}, Moving Average (MA) \cite{box1970distribution}, and Auto-Regressive (AR) \cite{shibata1976selection} are three basic mathematical models used in time series prediction. Some researchers combine them to obtain a general Auto-Regressive Moving Average (ARMA) model \cite{hamilton1994time}. Further, some variations like the ARIMA \cite{contreras2003arima} and Seasonal ARIMA \cite{szeto2009multivariate} were proposed to explain the intrinsic correlation of the sequence to predict the future. In financial applications, Generalized AutoRegressive Conditional Heteroskedasticity (GARCH) \cite{bauwens2006multivariate} is a common algorithm, which takes into account the volatility clustering of financial sequence. Others like the Vector Auto-Regression model (VAR) \cite{juselius2006cointegrated}, grey model \cite{tang2015based} and Hawkes Point Process \cite{hawkes1971spectra, ouyang2018modeling} also play their role in prediction. By establishing a mathematical model to fit the historical time trend curve, they predict the future only based on the development trend of history. Therefore, they only work well when the time series shows a clear trend or seasonal behavior.

\subsubsection{Statistical machine learning methods}
Some of the well-known regressive methods in machine learning are not suitable for time series prediction, such as linear regression, stepwise regression, ridge regression \cite{shi2016bayesian}, and lasso regression \cite{li2014forecasting}. The reason is that in the regression analysis, we assume that the data are independent of each other. In other words, the order of data can be swapped at will. However, in time series prediction, there is a temporal correlation between the data, and the time step is a very important cue for prediction. Others like SVR \cite{bao2014multi}, random forest \cite{liaw2002classification}, and Xgboost \cite{chen2016xgboost} can be used for the sequence prediction. However, they do not consider the complex coupling relations between multiple influencing factors \cite{cao2011coupled, cao2015coupling}. Accordingly, they are unable to cope with complex situations and cannot model the couplings of multiple factors and their influence on delivery dynamics.

\subsubsection{Deep learning methods}
 Recently, deep learning approaches \cite{yu2020predicting, zhang2019deep, zhu2020deep, zhang2020semi, sun2020dual, lin2020preserving, li2020autost} have been successfully applied to many sequence prediction scenarios \cite{zheng2014urban}, such as traffic prediction \cite{yu2017spatio, guo2019attention, diao2019dynamic, pan2019urban, wang2020traffic}, crime prediction \cite{yi2018integrated}, human mobility prediction \cite{zhang2017deep, feng2018deepmove, bai2019stg2seq}, and bike-sharing systems re-balancing \cite{li2019citywide, wang2018bravo, ouyang2018competitivebike, geng2019spatiotemporal}, etc. Attention based Seq-to-Seq model \cite{bahdanau2014neural} was first proposed to the field of Natural Language Processing (NLP). Further, in the time series prediction field, DA-RNN \cite{qin2017dual} applied a dual-stage attention mechanism to adaptively selects the relevant features and captures temporal dependencies. DeepTTE \cite{wang2018will} presents an attribute component that integrates external factors with the raw GPS sequence. However, both of the above methods cannot be directly used in this problem as lack of handle to heterogeneous features. GeoMan \cite{liang2018geoman} proposed a multi-level attention mechanism to model the dynamic Spatio-temporal correlations and fused the effects of external factors. But it failed to capture the accumulative and dynamic effects of features on the correlated sequence. Besides, its spatial attention is not suitable for this problem setting. In summary, existing research has made a lot of effort on analyzing sequence temporal dependency and optimizing models, yet rarely notice heterogeneous and coupling relationships \cite{ijcai_ZhangCZLS18,  cao2010depth, JianPCLG19} among sequence-feature pairs.

\begin{table}[!t]
\caption{Notations used in this paper \label{tab:one}}
\centering
\begin{tabular}{cl}
\hline\hline
\textbf{Symbol}              & \textbf{Definition}                                                                                                                              \\ \hline
$X$                          & \begin{tabular}[c]{@{}l@{}}A series of heterogeneous features associated with express delivery sequence\end{tabular} 
 
  \\ \hline
$x_{t}^{i}$                  & $i$-th feature in feature vector at time $t$ 
 
 \\ \hline
$\tilde{x}_{t}$     & $n$-dimensional vector of features at time $t$                                                                                                    \\ \hline
$\tilde{\mathbf{x}}_{t}$                  & All of the selected features within a period of time that corresponds to time $t$                                                                                                    \\ \hline
$Y$                          & Historical express delivery sequence, numerical time series of model input                                                                       \\ \hline
$y_{t} , \hat{y_{t}}$                      & The real number and predicted result of daily express parcels of the target region at time $t$    

 \\ \hline
$E$                          & Predicted result of express delivery sequence  

\\ \hline
$k$                          & The length of predicted express delivery sequence                                                                                                \\ \hline
$h$                          & The length of historical express delivery sequence                                                                                               \\ \hline
$l$                          & The range of features' effect, where the length of feature time window is $\left(2l+1 \right)$                                                                 \\ \hline
$n , n^{\prime}$                          & Dimension of features as well as their hidden representations                                                                       \\ \hline
$\alpha , \beta$                     &  Weight matrices of temporal attention and feature attention                                                                               \\ \hline
$h_{t}$                      & The output of encoder hidden state of express delivery sequence at time $t$                                                                      \\ \hline
$d_{t}$                      & The output of heterogeneous feature representation at time $t$                                                                                                   \\ \hline
$z_{t}$                      & \begin{tabular}[c]{@{}l@{}}A connection parameter to obtain a comprehensive representation from multiple hidden states\end{tabular}           
\\ \hline
$s_{t}$                      & The output of decoder hidden state

\\ \hline
$c_{t}^{H} , c_{t}^{D}$                  & Weight-redistributed context vector of encoder hidden state / feature hidden representation                                                                                       \\ \hline
$\varphi(\cdot) , \psi(\cdot)$             & The embedding method for numerical and categorical features respectively 

\\ \hline
g(.)                         & The LSTM functions for encoder and decoder

\\ \hline
$W_{*}, U_{*}, V_{*}, b_{*}$ & Learnable parameter matrix                                                                                                                       \\ \hline
\end{tabular}
\end{table}

\section{Problem Definition and Challenges}
\label{sec:PD}
\subsection{Research Problem}
This section first presents the definition of the express delivery sequence and feature representation, and then provides formalization of the studied problem.

\textbf{Definition 3.1.} Express Delivery Sequence: The historical express delivery sequence can be treated as a numerical time series $Y=\left(y_{t-h+1}, y_{t-h+2}, \cdots, y_{t}\right)$, where $y_{t}$ denotes the number of daily express parcels of the target region at time $t$, and $h$ is the length of historical sequence. Correspondingly, the predicted result can be formalized as $\left(\hat{y_{t+1}}, \hat{y_{t+2}}, \cdots, \hat{y_{t+k}}\right)$, where $k$ represents the length of the predicted sequence.

\textbf{Definition 3.2.} Heterogeneous Feature: There are 4 types of features used in the model, including temperature, weather, holiday and week. At time $t$, the $i$-th feature can be formulated as $x_{t}^{i}$. And all the features can be denoted as $\tilde{x}_{t}=\left(x_{t}^{1}, x_{t}^{2}, \cdots, x_{t}^{n}\right)$, where $n$ represents the type of the feature. Considering the cumulative effect, we have selected all the features within a period of time that corresponds to time $t$. It can be organized as $\tilde{\mathbf{x}}_{t}=\left(\tilde{x}_{t+1-l}, \cdots, \tilde{x}_{t+1}, \cdots, \tilde{x}_{t+1+l}\right)$, and $l$ denotes the length of feature time window.

\textbf{Problem.} Express Delivery Prediction: Given two types of input: $Y$ and $X$, which represent historical express delivery sequence and heterogeneous features, respectively. The goal of express delivery sequence prediction is to predict the future result over the next $k$ days for the target region. Specifically, the research problem is defined as follow:

\begin{equation}
\begin{array}{l}
\hat{{y}_{t+1}}= F\left(y_{t-h+1}, y_{t-h+2}, \cdots, y_{t} ; \tilde{\mathbf{x}}_{t}\right) = F\left(y_{t-h+1}, y_{t-h+2}, \cdots, y_{t} ; 
\tilde{x}_{t+1-l}, \cdots, \tilde{x}_{t+1}, \cdots, \tilde{x}_{t+1+l}\right) \\
E=\left\{y_{t+1}, y_{t+2}, \cdots, y_{t+k}\right\}
\end{array}
\end{equation}

Function $F$ is the proposed model, DeepExpree, and $\hat{{y}_{t+1}}$ denotes the predicted result as time $t+1$. To predict the next time result, $\hat{{y}_{t+1}}$ is joined with historical express delivery sequence as well as correlated features at time $t+2$. The prediction process is repeated $k$ times via previously predicted results to generate the sequence $E$. And all symbols appeared in the paper are summarized in Table \ref{tab:one}.  

\begin{figure}[!t]
    \centering
    \subfigure[]{
        \label{fig:3.1a} 
        \includegraphics[width=0.32\textwidth]{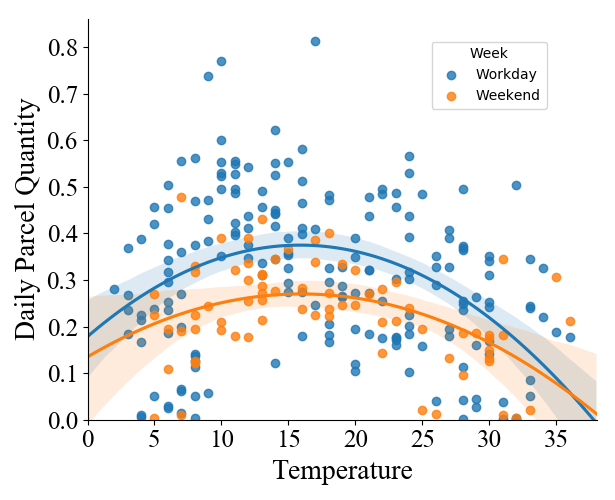}
    }%
    \subfigure[]{
        \label{fig:3.1b} 
        \includegraphics[width=0.32\textwidth]{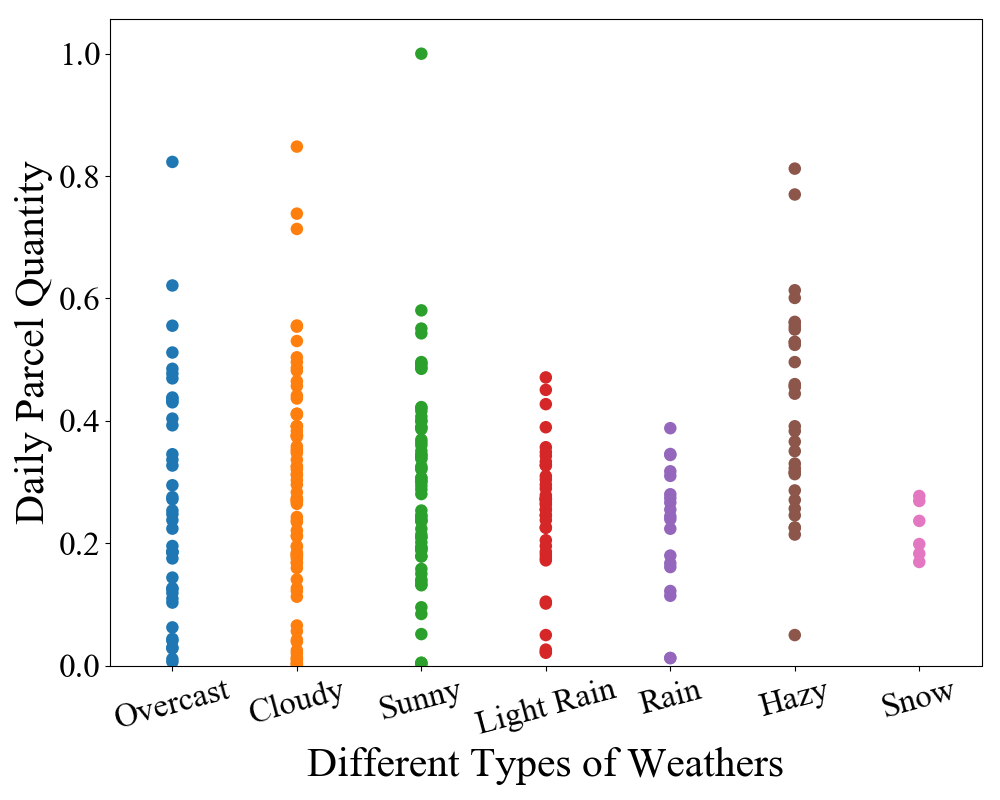}
    }%
    \subfigure[]{
        \label{fig:3.1c} 
        \includegraphics[width=0.32\textwidth]{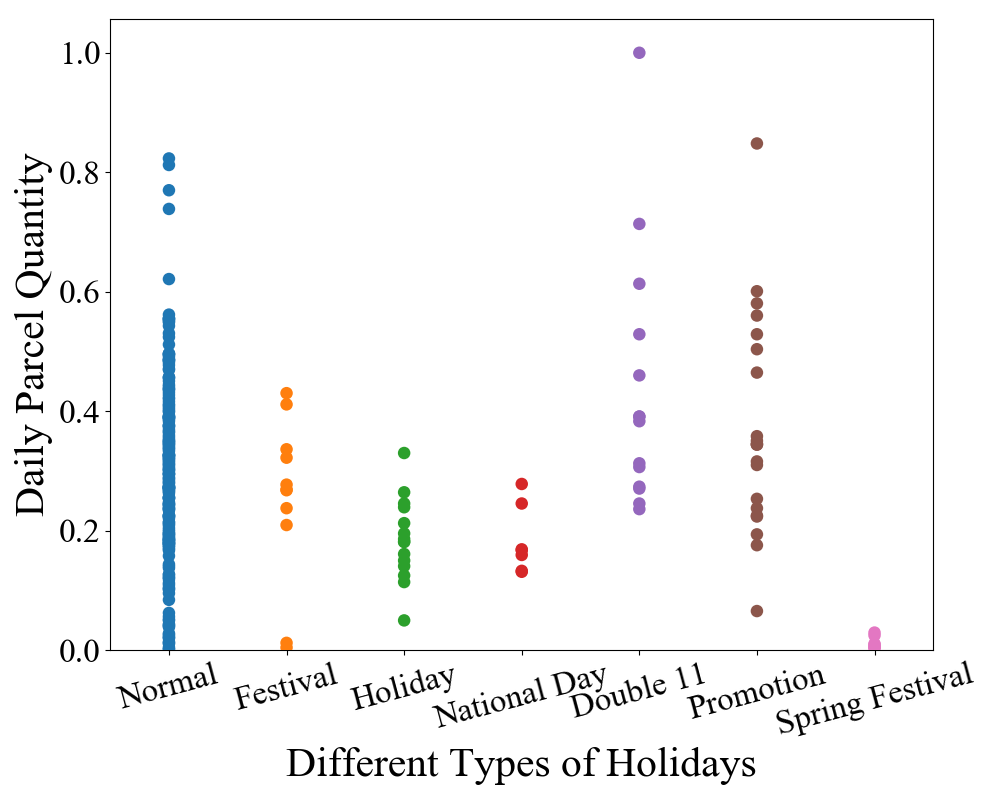}
    }%
    \centering
\caption{Relationships between express delivery and different features. (a) fit the relationship between temperature and daily parcel quantity in both workday and weekend. (b) and (c) show the distribution of daily parcel quantity for each type of weather and holiday respectively.}
\label{fig:3}
\end{figure}

\subsection{Data Heterogeneity Situation}
\label{DHS}
For accurate prediction and information preservation, one thing challenging yet crucial to this problem is the heterogeneous situations of the feature, which illustrate the differences between different data. The definition of heterogeneous includes many aspects, e.g. different data sources, data types, distributions, ranges, couplings, and so on. Specifically, in the DeepExpress model, we used one time series and four features as model input respectively. Express delivery sequence is a numerical time series that records the daily parcel quantity of the target region. Therefore, its value has a clear meaning. That is, the larger the value, the more parcels. To showcase the relationship between temperature feature and delivery sequence, we present the daily parcel quantity and temperature of one university in Xi’an city in Fig \ref{fig:3.1a}. From the figure, we can observe that although the temperature is numerical data, it has a different meaning from the value of express delivery. With the increase of temperature, the daily parcel quantity changes in a quadratic curve. Also, their relationship even stronger at weekends, because people have to go out for work or study during the workday, whether it’s cold or hot. Therefore, the predictive model should be able to learn heterogeneous feature week and temperature together as well as capture the relation between express and temperature. Moreover, Fig \ref{fig:3.1b} and Fig \ref{fig:3.1c} show the daily parcel quantity with different categorical features, the figure illustrates that the distribution of different features is uneven and shows different relations. That is the features with relative more frequency (a more concentrated scatter distribution) have little effect, while the less frequent features have a greater effect. Valuable information will be lost when these features are directly embedded and treated equally, which might hinder the prediction accuracy. Therefore, we are required to consider an uneven distributed categorical feature, and adaptively map them to a hidden representation based on their relevance to the sequence. Finally, the express delivery sequence works on a long-term effects and related to the specific region of a city, while the heterogeneous features affect in short-term as well as city-level. So it's natural to model them separately. 

\subsection{Complex Coupling Situation}
Complex coupling between sequence and features is another challenge that worth noticing. Different from traditional time series (e.g. sensor reading, internet signal, and biomedical indicator, etc) that always have significant regularity, the act of sending and receiving parcels is one of people's behavior that suffers from more complicated impacts in various aspects. Therefore, the delivery sequence is not only has a strong correlation with its historical changes but also affected by the coupled external features. Two types of coupling relationships that appeared in this paper are introduced as follows:

\textbf{a) Accumulative effects}. The features' effect on the sequence is not always immediate, which could be accumulated for a period of time. Hysteresis is one of the accumulative effects that appeared in this problem. The feature change in the previous period will affect the sequence in the next period of time. For example, bad weather (e.g., snowy and rainy) leads to some traffic problems like traffic jams and highway closed, which will cause delay packages delivery. And also, stop services of express during holidays leading to the overstock of parcels, and further results in the growing demand after the holiday. These impacts will take effect gradually rather than immediately. The pre-impact is another situation that current changes in the sequence will be affected by future features. The reason is that people will make decisions based on future events. For instance, people often make preparations through the weather forecast, e.g., buying masks before the hazy weather. Besides, online promotions always start before festivals and will last for a while. Hence the aforementioned pre-determination will lead to the result that the delivery sequence is coupled with future features.

\textbf{b) Dynamic changes}. The relationship between the delivery sequence and features is not stable, features at different times also have different effects on the sequence and they change dynamically over time. For example, bad weather will have a negative impact on express delivery, while promotions have the opposite effect. When they appear at the same time, the model needs to weight features based on their impacts as well as time intervals. And the weighted value will be changed as the feature’s value and intervals changed. However, it is quite challenging to learn the correlation between the delivery sequence and features from dynamic changes and assign weights adaptively, which requires a joint training mechanism.

\begin{figure}[!t]
	\centering
	\includegraphics[width=0.5\linewidth]{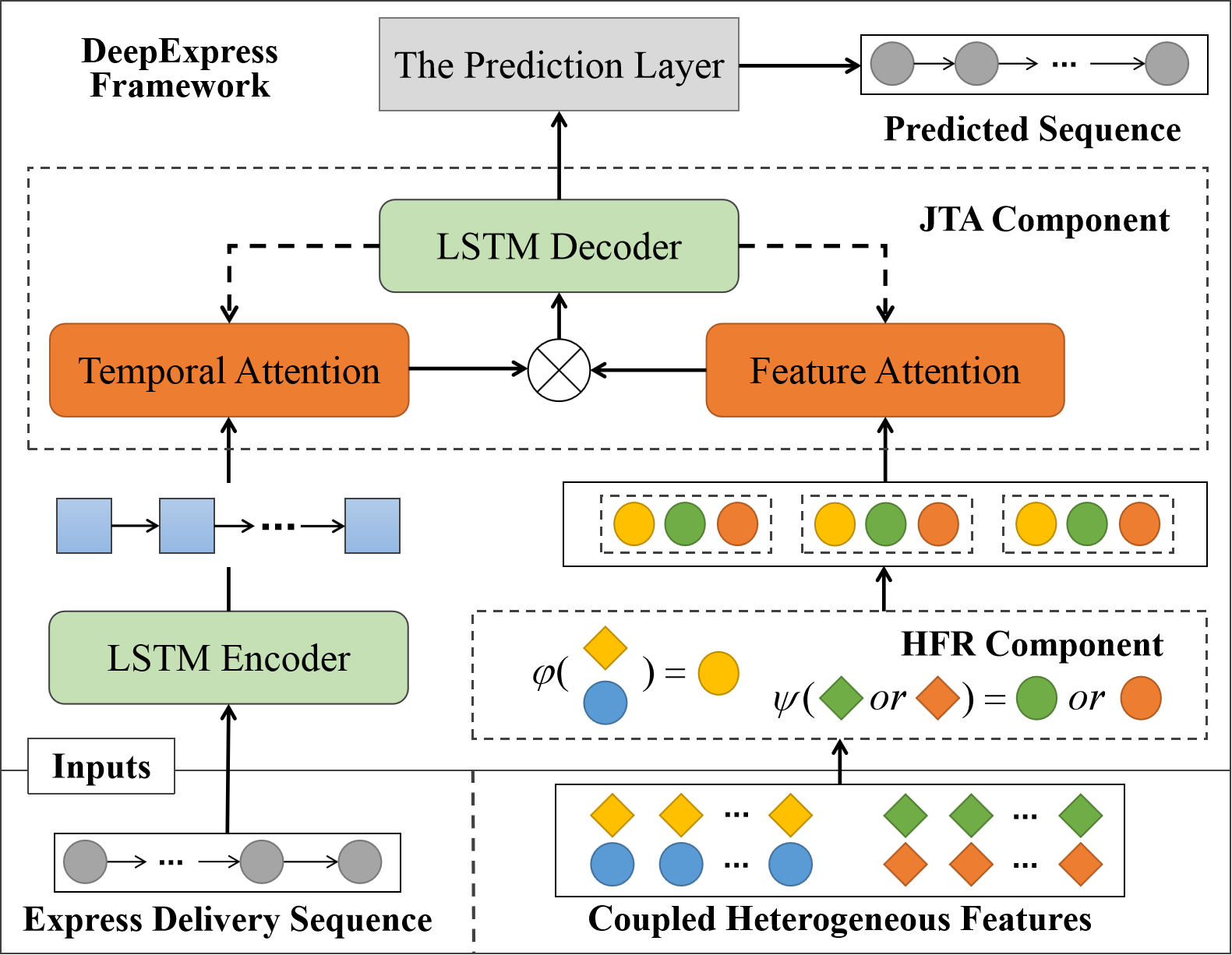}
	\caption{Framework of the DeepExpress. An express delivery seq2seq learning component to model the auto-regressive process of delivery sequence. Heterogeneous feature representation component is used to process different types of heterogeneous urban data. Joint training attention mechanism designed to learn complex couplings.}
	\label{fig:framework}
\end{figure}

\section{Methodology}
\label{sec:MD}
This section first presents the framework of the proposed model and then introduces the components step by step. The framework of the DeepExpress is shown in Fig. \ref{fig:framework}, which consists of three components: express delivery seq2seq learning, heterogeneous feature representation (HFR), and joint training attention mechanism (JTA). The first component aims to combine features' impact with delivery sequence auto-regressive process. This component first takes the RNNs-based encoder-decoder architecture to model the auto-regressive process of delivery sequence and selects various features in a time window as additional inputs. Then the features are fused in the decoder to enhance performance. The second component is designed to process heterogeneous data, which first process the different features separately, then learn the mapping algorithm between features and sequences, and finally fuse all of features into a 1D hidden representation. The third component is dedicated to solving the problem of complex coupling between the delivery sequence and features. It combines the temporal attention of the sequence with feature attention through a joint training attention mechanism, which could improve the final prediction performance by considering the complex couplings, i.e. accumulative effects as well as dynamic changes.

\subsection{Express Delivery Seq2seq Learning}
\label{sec:ESAL}
For the studied problem, the delivery sequence changes as a joint effect of sequence auto-regression and features’ effect. The underlying intuition is that extend the classic seq2seq framework to combine the external features' effect and sequence auto-regressive in the prediction model. Specifically, the attention-based encoder-decoder network is a state-of-art seq2seq approach that used to capture the long-term temporal dependencies of the sequence. So the express delivery sequence is used as one part of the model input, and their auto-regression process can be learned by an attention-based seq2seq model. And also, considering the hysteresis and pre-impact situation of accumulative effects, the external features of the period before and after also need to be considered. Therefore, features within a fixed time window are employed as another part of the input. And the features' effect can be fused in the decoding stage of the model.  

\textbf{Encoder with Temporal Attention.} Given a series of historical parcel volumes $Y=\left(y_{t-h+1}, y_{t-h+2}, \cdots, y_{t}\right)$, the encoded representations can be computed with $h_{t}=g\left(h_{t-1}, y_{t}\right)$, where $g(\cdot)$ represents a LSTM function, and $h_{t}$ is the output of encoder hidden state of express delivery sequence at time $t$. In order to capture the long-term temporal dependencies of the sequence, a temporal attention is used in the encoder stage, and it can be formulated as below:

\begin{equation}
\label{eq3}
\alpha_{t+1}^{i}=\operatorname{softmax}\left(V_{T} \tanh \left(W_{T} s_{t}+U_{T} h^{i}+b_{T}\right)\right), t-h+1 \leq i \leq t
\end{equation}

$V_{T}$, $W_{T}$, $U_{T}$, $h_{T}$ are learnable parameters. Specifically, $s_{t}$ is the previous hidden state of the decoder which contains comprehensive information from a set of previous hidden states, and $\alpha$ is an attention weight matrix that calculates the relevance between $s_{t}$ and each hidden state in the sequence (from time $t-h+1$ to time $t$). By doing this, the temporal attention can learn the temporal couplings between the input sequence and the whole model. Besides, a softmax function is used to make sure the sum of all the attention weights equal to 1.

\begin{equation}
\label{eq4}
c_{t}^{H}=\sum_{i=t+1-h}^{t} \alpha^{i} h^{i}
\end{equation}

$c_{t}^{H}$ is a weighted context vector that sums all hidden states of the sequence and their corresponding attention weights at time $t$, by which the decoder can access the entire input sequence and focus on the relevant time step in the delivery sequence.

\textbf{Decoder with Features’ Effect.} Express delivery sequence has a strong correlation with the external features, e.g. weather, holiday, and temperature. We use $l$ to represent the range of features' accumulative effect, and employed all the features before and after time $t$ with in the range $l$. Thus the length of feature time window is ($2l+1$), and the features input at time $t$ can be formulated as $\tilde{\mathbf{x}}_{t}=\left(\tilde{x}_{t+1-l}, \cdots, \tilde{x}_{t+1}, \cdots, \tilde{x}_{t+1+l}\right)$. Input features are transformed by means of the min-max scaler and embedding methods separately. They are uniformly expressed as $d_{t}=e\left(\widetilde{\mathrm{x}}_{t}\right)$, where $e(\cdot)$ is denoted as embedding method. To predict the next time result, they are formalized as a 1D vector and fused with weighted context vector of the sequence through $z_{t}=\left[c_{t}^{H} ; d_{t}\right]$ in the decoder stage. It aims to decode the concatenated information to make final prediction. The concatenated hidden states are decoded with the LSTM units in Eq. \ref{prediction}:

\begin{equation}
\label{prediction}
s_{t+1}=g\left(z_{t+1} ; s_{t}\right)
\end{equation}

And then a Multi-Layer Perception (MLP) is employed as the prediction layer to estimate the future result, which is represented as $f(\cdot)$. And the predicted result of $\hat{y}_{t+1}$ can be computed by:

\begin{equation}
\begin{aligned}
\hat{y}_{t+1} &=f\left(s_{t+1}\right) &=\operatorname{soft} \max \left(W_{P}^{2}\left(\sigma\left(W_{P}^{1} s_{t+1}+b_{P}^{1}\right)\right)+b_{P}^{2}\right)
\end{aligned}
\end{equation}

The matrix $W_{P}^{*}$ and the vector $b_{P}^{*}$ are the learnable parameters of the prediction function, and $\sigma$ is the sigmoid function. In this way, the final prediction of future express delivery can be conducted.

\subsection{Heterogeneous Feature Representation}
\label{sec:HFR}
As shown in Fig. \ref{fig:framework}, the model contains two types of inputs: historical express delivery sequence and their correlated external features. As mentioned in section \ref{DHS}, we have found out that the features are heterogeneous in terms of data source, type, distribution, range, and impact with the sequence. Most of the existing time series prediction methods omit the heterogeneous situations of the features, and directly encoding them into a network. Specifically, the numerical features are normalized and directly input without any transformation, and the categorical features are simply transformed into a low-dimensional vector by feeding them into embedding layers \cite{liang2018geoman}. The express delivery sequence is a numerical time series, and its value has specific meanings, while the above methods do not consider the correlation between different features and the delivery sequence. The heterogeneous features are transformed and mapped to the same and independent representation, which will cause information loss and wrong representation problem.

In order to leverage the useful information of data as much as possible, this paper designs a novel heterogeneous feature representation. It first assigns appropriate embedding methods to different features, and then integrates and trains them with the entire model. By doing this, the parameters of the embedding methods are updated with the overall model training process, and the hidden representation could contain both the information of the feature itself as well as the relationship between the feature and the delivery sequence. Algorithm \ref{alg:1} describes the representation process for heterogeneous features.

\begin{algorithm}
\caption{Heterogeneous Feature Representation}
\label{alg:1}
\renewcommand{\algorithmicrequire}{\textbf{Input:}}
\renewcommand{\algorithmicensure}{\textbf{Output:}}
\begin{algorithmic}[1]
\REQUIRE ~~\\ 
Heterogeneous features $X$; \\
Express delivery sequence $Y$;
\ENSURE ~~\\ 
Feature hidden representation $d$;
\STATE $//$ Generate representation;
\FOR{$i=1$ to $n$}
    \FOR{$j=t+1$ to $t+k$}
        \IF{$x$ is numerical feature} 
            \STATE $d \Leftarrow \varphi^{i}\left(x_{j}^{i}\right) , \forall x_{j}^{i} \in X$
        \ELSE 
            \STATE $d \Leftarrow \psi^{i}\left(x_{j}^{i}\right) , \forall x_{j}^{i} \in X$
        \ENDIF 
    \ENDFOR
\ENDFOR
\STATE $//$ Training the model;
\STATE initialize all learnable parameters in DeepExpress
\REPEAT
\STATE generate hidden representation $d$ from $X$
\STATE training the overall model with data $(d,Y)$
\UNTIL{model overfitting} 
\end{algorithmic}
\end{algorithm}

Specifically, function $\varphi(\cdot)$ is an embedding method for numerical feature (i.e. temperature). Based on the data analysis result in section \ref{sec:PD}, we notice that temperature has strong yet different relations with express delivery on weekend and workday. On weekends, there is an obvious conic relationship between express delivery and temperature, and we multiply two neural networks to model this relation. Because the expansion of its formula is a one-variable quadratic equation. But on workdays, the relationship is too complicated to be observed directly, so we use an MLP to learn this complex nonlinear relationship. Therefore, the embedding method for temperature and week feature can be represented as:

\begin{equation}
\varphi(x)=\left\{\begin{array}{c}
(W \cdot x+b) \otimes(V \cdot x+c), \text { if week } \in \text { weekend } \\
f(x), \text { if week } \in \text { workday }
\end{array}\right.
\end{equation}

$W$, $V$, $b$, $c$, are learnable parameters that trained together with the model, and $f\left(x\right)$ is denoted as a MLP function. 

Besides, function $\psi(\cdot)$ is an embedding method designed for categorical features (i.e. weather and holiday). It first converts the categorical feature into 0-1 values through one-hot encoding and then multiplies them with a learnable parameter vector. By updating the parameters, it can learn the relationship between the delivery sequence and each category. Moreover, each categorical feature requires using different embedding methods as they own different distribution and correlations.

\begin{figure}[!t]
	\centering
	\includegraphics[width=0.8\linewidth]{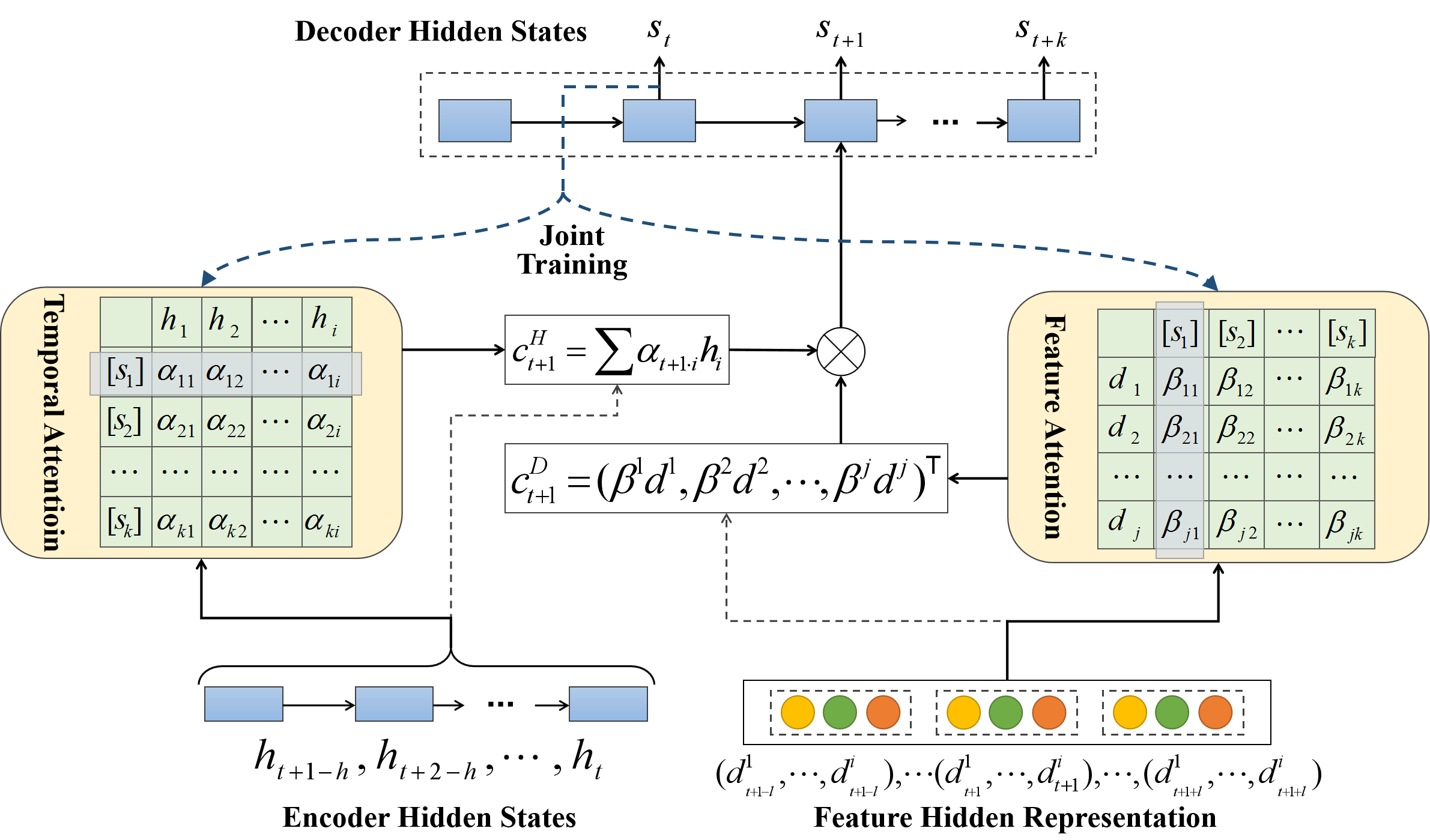}
	\caption{Joint Training Attention Mechanism of the DeepExpress.}
	\label{fig:mechanism}
\end{figure}

\subsection{Joint Training Attention Mechanism}
\label{sec:JTAM}
One important obstacle for achieve accurate express delivery sequence prediction is the complex couplings situation between the delivery sequence and features. Previous methods assume that variables are independent with each other for simplify, and trained them separately and fused all of the result in the end. These methods omit associations between variables, thus they can neither learn the mutual affects between each other nor dynamically assign weights according to changes in features.  Although in recent years, bunch of works on multi-variate time series have been proposed \cite{qin2017dual, salinas2019high, li2019ea, yuan2018muvan, hu2019transformation, wan2019multivariate, shen2018novel}, the proposed methods mainly focus on select the most influenced variables, which weaken the heterogeneous situation of data and trained the target sequence with external features together. Hence, selecting features with the same length as the sequence will cause data redundancy, which result in unsatisfactory prediction performance.

To model the complex coupling relationship between sequence and features and also to assign weights to most relevant features adaptively, a joint training attention mechanism is provided here. Specifically, we first replace the function $e(\cdot)$ in Sec. \ref{sec:HFR} with feature attention, it is used to calculate the relevance between the features’ hidden representations and decoder hidden states. Both of the weighted context vectors of sequence temporal attention as well as feature attention are then concatenated as the input of the decoder. By doing this, the two attention mechanisms could be jointly trained through back-propagation. Furthermore, the model can easily learn long-term temporal dependencies and short-term features couplings at the same time. 

\textbf{Feature Attention.} The original feature inputs within the time window are first transformed into a hidden representation through the heterogeneous feature representation component, which could be formulated as $\left(d_{t+1-l}^{1}, \cdots d_{t+1-l}^{i}, d_{t+2-l}^{1}, \cdots d_{t+1+l}^{i}\right), i \in n^{\prime}$ and $n^{\prime}$ is the dimension of hidden representations. Similar with the temporal attention in encoder, the equation of feature attention can be formulated as below:

\begin{equation}
\label{eq5}
\beta_{t}^{j}=\operatorname{softmax}\left(V_{F} \tanh \left(W_{F} s_{t-1}+U_{F} d_{t}^{j}+b_{F}\right)\right), 1 \leq j \leq n^{\prime}\left(2l+1\right)
\end{equation}

All of the hidden representations within the $\left(2l+1\right)$ time window are organized into 1D vector, thus the length of vector is $n^{\prime}\left(2l+1\right)$. $V_{F}$, $W_{F}$, $U_{F}$, $h_{F}$ are learnable parameters, and $\beta$ is an attention weight matrix that calculates the relevance between decoder hidden states $s_{t}$ and each of hidden representation in the vector. In this way, the feature attention can capture the feature couplings between the features and the entire model. After that, the matrix $\beta$ is normalized with a softmax function.

The calculated attention weights are multiplied with their corresponding hidden representations to select the most relevant features and generate the weighted context vectors of feature attention, the output of the feature attention can be formulated as follows:

\begin{equation}
c_{t}^{D}=\left(\beta_{t}^{1} d_{t}^{1}, \beta_{t}^{2} d_{t}^{2}, \cdots, \beta_{t}^{n^{\prime}} d_{t}^{n^{\prime}}\right)^{\top}.
\end{equation}

$\mathcal{C}_{t}^{D}$ is a weights-redistributed vector where the features dynamically assign weights according to their impacts. By doing this, the model can adaptively select the most influential features to the express delivery.
  
\textbf{Decoder with Joint Training Attention.} Specifically, we combine the hidden state from feature attention and temporal attention to generate a comprehensive representation for the decoder, and the fusion function in Sec. \ref{sec:ESAL} is updated in Eq. \ref{eq2}. 

\begin{equation}
\label{eq2}
z_{t+1}=\left[c_{t+1}^{H} \otimes c_{t+1}^{D}\right]
\end{equation}

$z_{t+1}$ is a parameter that contains joint information from: $c_{t+1}^{H}$ and $c_{t+1}^{D}$. $c_{t+1}^{H}$ is the weighted context vector for delivery sequence. It contains long-term information about historical sequences which is used to predict the result without the effects of external features. $c_{t+1}^{D}$ is the weighted context vector for feature representation. It is a comprehensive representation of the external feature effects that indicate the degree of feature effects on the sequence (equivalent to the coefficient). And the symbol $\otimes$ denotes the inner product of these two vectors. It serves as the next-step input of the decoder for prediction. By doing this, the output of decoder hidden state $s_{t}$ in Eq. \ref{eq4} could contain information from both delivery sequence as well as coupled features. In each attention mechanism, the correlation between $s_{t}$ and hidden states $h^{i}$, $d_{t}^{j}$ are calculated through the match function, and the relevance $\alpha_{t}^{i}$, $\beta_{t}^{j}$ between each variable and the output sequence can be obtained. In this way, two mechanisms are jointly trained within one feed-forward neural network and model the couplings between multiple factors simultaneously.  

\subsection{Model Training}
The integration of multiple components within one single model can be effectively achieved by a feed-forward neural network. Since the model is smooth and differentiable, it can be jointly trained using standard back-propagation. Specifically, we design a grid-search algorithm to automatically set hyper-parameters, including the number of units, batch size, epochs, time steps, and so on. The loss function of the model is the Mean Absolute Error (MAE) which is defined in Eq. \ref{loss}. $N$ is the number of training samples, $\hat{y}_{t+1}$ denotes the predicted result and $y_{t+1}$ represents the real record. 

\begin{equation}
\label{loss}
\text { loss }=\frac{1}{N} \Sigma_{i=1}^{N}\left(\hat{y}_{t+1}^{i}-y_{t+1}^{i}\right)^{2}
\end{equation}

\section{Evaluation}
\label{sec:EV}
In this section, we conduct extensive experiments based on real-world express delivery datasets to answer the following questions:

\textbf{Q1:} Does our proposed DeepExpress outperforms existing methods in express delivery prediction?

\textbf{Q2:} What hyper-parameters of DeepExpress are important to tune the model?

\textbf{Q3:} Weather the proposed joint training attention mechanism successfully captures the complex couplings between the sequence and features? 

\begin{table}[!t]
\caption{Details of the Experimental Dataset \label{tab:two}}
\centering
\begin{tabular}{c|l|l}
\hline\hline
\multicolumn{2}{c|}{\textbf{Dataset}}       & \multicolumn{1}{c}{\textbf{Description}}                                           \\ \hline
\multirow{5}{*}{\begin{tabular}[c]{@{}c@{}}Express \\ Delivery\end{tabular}} &
  Target series &
  \begin{tabular}[c]{@{}l@{}}The quantity of daily sent or received parcels\end{tabular} \\
 &
  Region types &
  \begin{tabular}[c]{@{}l@{}}4: \{University, Company, Urban village, Residential\}\end{tabular} \\
                         & Time spans       & \begin{tabular}[c]{@{}l@{}}Jun. 2016-Jun. 2017, Nov. 2017-Sep. 2018\end{tabular} \\
                         & Time intervals   & 1 day                                                                              \\
                         & \# Instance       & 5,881,166                                                                          \\
\multirow{2}{*}{\begin{tabular}[c]{@{}c@{}}Weather \\ Forecast\end{tabular}} &
  \begin{tabular}[c]{@{}l@{}}\# Weather conditions\end{tabular} &
  15 \\
                         & Temperature/$^{\circ} \mathrm{C}$ & -5$\sim$40                                                                         \\
\multirow{2}{*}{Holiday} & \#Categories     & 4                                                                                  \\
                         & \#Label          & 7                                                                                  \\ \hline
\end{tabular}
\end{table}

\subsection{Datasets}
\label{dataset}
The proposed model was evaluated based on a real-world large scale dataset which contains three sub-datasets: the historical express delivery dataset, holiday schedule dataset, and weather forecast dataset, as depicted in Table \ref{tab:two}. We set 60\% of the dataset as training samples, 20\% as testing samples, and the remaining 20\% for validation.

\subsubsection{Express delivery data}
The express sheets recorded by an express logistics company provide the real-world express information of a Chinese city. By collecting the express sheets of several express logistics companies in Xi'an for the latest two years, we are able to obtain most of the records of the city. Specifically, we take two types of delivery sequence (i.e. sending and receiving parcels) within four types of regions (i.e. university, company, urban village, and residential area) as the express delivery dataset, which could cover the main regional types and delivery behaviors of a city. And also, the data that belongs to the university region contains samples from 20 universities of the city, which is the largest subset of the datasets and more representative. Therefore, we use it as the test dataset for the ablation study.

\subsubsection{Holiday schedule data}
We collect the public information like China's legal holiday schedule, Chinese festival calendar and the schedule of e-commerce activities, and categorize them into four types: e-commerce activity day (Ali New Year goods festival, JD 618, etc.), festival (Father's Day, Chinese Valentine's Day, etc.), statutory holiday (Dragon Boat Festival, Mid-Autumn Festival, etc.), and some special moments (Double 11, National Day and Spring Festival). Then, we label all the time periods which belong to the four categories, and the rest of time periods are not considered in this dataset and labeled as ordinary time.

\subsubsection{Weather forecast data}
We collect historical weather forecast data of Xi'an from the China weather forecast website. Each weather forecast record consists of four items: daytime weather condition, daytime temperature, nighttime weather condition, and nighttime temperature. As people usually send and receive parcels at the daytime, we only take daytime into consideration. Besides, we classify all the weather conditions into fifteen types (sunny, cloudy, hazy, cloudy, and snow, etc.) and convert each original daytime weather condition to one of them.

\subsection{Baselines}
We compare the proposed model with several baseline models, which are the most typical of temporal prediction methods, as mentioned in section \ref{sec:RW}, including:

\textbf{SARIMA \cite{szeto2009multivariate}:} ARIMA is one of the most general statistical models for forecasting a time series and separating the signal from the noise. SARIMA extends it to learn both closeness and periodic dependencies through a seasonal component.

\textbf{Prophet \cite{taylor2018forecasting}:} Prophet is a practical time series prediction approach developed by Facebook, it aims at forecasting 'at scale' by combining configurable models with analyst-in-the-loop performance analysis. 

\textbf{SVR \cite{bao2014multi}:} SVR is a classic support vector regression model. It is an important prediction model solve a variety of nonlinear regression problems without deep learning.

\textbf{Xgboost \cite{chen2016xgboost}:} Xgboost is one of Gradient Boosting Decision Tree (GBDT). It takes an ensemble method to fit current residuals and gradients of the loss function in a forward step-wise manner, which can be used for almost all of the linear and nonlinear regression problems.

\textbf{Seq2seq \cite{sutskever2014sequence}:} Seq2seq is an encoder-decoder based model that map inputs to a fixed vector with LSTM encoder, then feed the vector to another LSTM decoder for prediction.

\textbf{Att-Seq2seq \cite{liu2018context}:} Att-Seq2seq is an extension of the seq2seq model, which contains temporal attention mechanism to learn long-term temporal dependency of the sequence.

\textbf{DA-RNN \cite{qin2017dual}:} DA-RNN is a dual-stage attention based RNN network, which can adaptively select the most relevant features and capture the long-term temporal dependencies appropriately.

\textbf{EA-LSTM \cite{li2019ea}:} EA-LSTM is a hybrid model that combines evolutionary attention-based LSTM network with competitive random search and a collaborative training mechanism.

\subsection{Modeling Details}
\textbf{Model Inputs and Setting.}
Since conventional models including SARIMA, Prophet, SVR and Xgboost are not supported to input the sequence and discrete features at the same time (features can be processed to concatenate with sequence, but redundant information causes a sharp decline of the model performance), we only take express delivery sequence data as a single dimension input. And the rest of deep learning-based models, including Seq2seq, Att-Seq2seq, DA-RNN, EA-LSTM, as well as DeepExpress, are compared under the same experimental setting, by taking both sequence and features as model input. 

And we use the Sigmoid function as an activation function for the output in MLP. Employees a tanh function within the attention mechanism and a Relu as the activation function for the rest of fully connected layers. Besides, we apply Adam as an optimizer to train the model and set batch size as 64. Also, we use MAE as the loss function and MAE as well as ACC for compiling evaluate metrics. To prevent overfitting, we use a dropout function in the non-recurrent connection and set the epochs as 30 with the full training data. According to the experimental results, the model achieves the best performance when the hyperparameters are set to h=21, l=3, k=3. So we use these as the model setting in overall performance evaluation in section \ref{OP}. To be fair, all the deep learning-based models are trained five times and their average performance is calculated for comparison.

\textbf{Evaluation Metrics.}
To evaluate the performance of various methods for prediction, we consider two evaluation metrics, namely Root Mean Squared Error (RMSE) and Mean Absolute Error(MAE). Both of them are commonly used in evaluating regression tasks. Specifically, RMSE is the sum of the squared errors before taking the square root, which is used to measure the deviation between the predicted result and the real value. Assume that ${y}_{t}$ is the real value and $\hat{y}_{t}$ denotes the predicted result at time $t$, and $N$ is the total number of data samples, then the RMSE is defined in Eq. \eqref{eq:RMSE}.

\begin{equation}
RMSE=\sqrt{\frac{1}{N} \sum_{i=1}^{N}\left(y_{t}-\hat{y}_{t}\right)^{2}} \label{eq:RMSE}
\end{equation}

\begin{table}[!t]
\caption{Experimental Result with Different Regions and Different Delivery Behaviors. \label{tab:three}}
\centering
\begin{tabular}{cccccccc}
\hline
\multirow{2}{*}{\textbf{Dataset}}                                                          & \multirow{2}{*}{\textbf{Model}} & \multicolumn{2}{c}{\textbf{Metrics}} & \multirow{2}{*}{\textbf{Dataset}}                                                             & \multirow{2}{*}{\textbf{Model}} & \multicolumn{2}{c}{\textbf{Metrics}} \\ \cline{3-4} \cline{7-8} 
                                                                                           &                                 & \textbf{RMSE}     & \textbf{MAE}     &                                                                                               &                                 & \textbf{RMSE}     & \textbf{MAE}     \\ \hline
\multirow{9}{*}{\begin{tabular}[c]{@{}c@{}}University/\\ Send\\ Parcels\end{tabular}}      & SARIMA                          & 0.2669            & 0.2365           & \multirow{9}{*}{\begin{tabular}[c]{@{}c@{}}University/\\ Receive\\ Parcels\end{tabular}}      & SARIMA                          & 0.2837            & 0.2484           \\ \cline{2-4} \cline{6-8} 
                                                                                           & Prophet                         & 0.2428            & 0.2086           &                                                                                               & Prophet                         & 0.2229            & 0.1905           \\ \cline{2-4} \cline{6-8} 
                                                                                           & SVR                             & 0.2486            & 0.2093           &                                                                                               & SVR                             & 0.2417            & 0.2123           \\ \cline{2-4} \cline{6-8} 
                                                                                           & Xgboost                         & 0.2257            & 0.1934           &                                                                                               & Xgboost                         & 0.2017            & 0.1744           \\ \cline{2-4} \cline{6-8} 
                                                                                           & Seq2seq                         & 0.1537            & 0.1184           &                                                                                               & Seq2seq                         & 0.1763            & 0.1469           \\ \cline{2-4} \cline{6-8} 
                                                                                           & Att-Seq2seq                     & 0.1343            & 0.0992           &                                                                                               & Att-Seq2seq                     & 0.1583            & 0.1258           \\ \cline{2-4} \cline{6-8} 
                                                                                           & DA-RNN                          & 0.1337            & 0.0968           &                                                                                               & DA-RNN                          & 0.1526            & 0.1182           \\ \cline{2-4} \cline{6-8} 
                                                                                           & EA-LSTM                         & 0.1223            & 0.0871           &                                                                                               & EA-LSTM                         & 0.1467            & 0.1131           \\ \cline{2-4} \cline{6-8} 
                                                                                           & \textbf{DeepExpress}            & \textbf{0.1126}   & \textbf{0.0818}  &                                                                                               & \textbf{DeepExpress}            & \textbf{0.1259}   & \textbf{0.0917}  \\ \hline
\multirow{9}{*}{\begin{tabular}[c]{@{}c@{}}Company/\\ Send\\ Parcels\end{tabular}}         & SARIMA                          & 0.2663            & 0.2351           & \multirow{9}{*}{\begin{tabular}[c]{@{}c@{}}Company/\\ Receive\\ Parcels\end{tabular}}         & SARIMA                          & 0.2795            & 0.2472           \\ \cline{2-4} \cline{6-8} 
                                                                                           & Prophet                         & 0.2381            & 0.2058           &                                                                                               & Prophet                         & 0.2326            & 0.1961           \\ \cline{2-4} \cline{6-8} 
                                                                                           & SVR                             & 0.2418            & 0.2027           &                                                                                               & SVR                             & 0.2539            & 0.216            \\ \cline{2-4} \cline{6-8} 
                                                                                           & Xgboost                         & 0.2267            & 0.1893           &                                                                                               & Xgboost                         & 0.2176            & 0.1833           \\ \cline{2-4} \cline{6-8} 
                                                                                           & Seq2seq                         & 0.1594            & 0.1212           &                                                                                               & Seq2seq                         & 0.1749            & 0.1369           \\ \cline{2-4} \cline{6-8} 
                                                                                           & Att-Seq2seq                     & 0.1389            & 0.1027           &                                                                                               & Att-Seq2seq                     & 0.1494            & 0.1136           \\ \cline{2-4} \cline{6-8} 
                                                                                           & DA-RNN                          & 0.1336            & 0.1002           &                                                                                               & DA-RNN                          & 0.1474            & 0.1117           \\ \cline{2-4} \cline{6-8} 
                                                                                           & EA-LSTM                         & 0.1248            & 0.0905           &                                                                                               & EA-LSTM                         & 0.1327            & 0.1161           \\ \cline{2-4} \cline{6-8} 
                                                                                           & \textbf{DeepExpress}            & \textbf{0.1174}   & \textbf{0.0841}  &                                                                                               & \textbf{DeepExpress}            & \textbf{0.1228}   & \textbf{0.0872}  \\ \hline
\multirow{9}{*}{\begin{tabular}[c]{@{}c@{}}Urban\\ Village/\\ Send\\ Parcels\end{tabular}} & SARIMA                          & 0.2328            & 0.1964           & \multirow{9}{*}{\begin{tabular}[c]{@{}c@{}}Urban\\ Village/\\ Receive\\ Parcels\end{tabular}} & SARIMA                          & 0.2901            & 0.2586           \\ \cline{2-4} \cline{6-8} 
                                                                                           & Prophet                         & 0.2136            & 0.1798           &                                                                                               & Prophet                         & 0.2304            & 0.1947           \\ \cline{2-4} \cline{6-8} 
                                                                                           & SVR                             & 0.2372            & 0.2052           &                                                                                               & SVR                             & 0.2659            & 0.2294           \\ \cline{2-4} \cline{6-8} 
                                                                                           & Xgboost                         & 0.1878            & 0.1511           &                                                                                               & Xgboost                         & 0.2207            & 0.1896           \\ \cline{2-4} \cline{6-8} 
                                                                                           & Seq2seq                         & 0.1518            & 0.1238           &                                                                                               & Seq2seq                         & 0.1964            & 0.1641           \\ \cline{2-4} \cline{6-8} 
                                                                                           & Att-Seq2seq                     & 0.1303            & 0.0958           &                                                                                               & Att-Seq2seq                     & 0.1714            & 0.1329           \\ \cline{2-4} \cline{6-8} 
                                                                                           & DA-RNN                          & 0.1385            & 0.1037           &                                                                                               & DA-RNN                          & 0.1735            & 0.1371           \\ \cline{2-4} \cline{6-8} 
                                                                                           & EA-LSTM                         & 0.1262            & 0.0933           &                                                                                               & EA-LSTM                         & 0.1672            & 0.1312           \\ \cline{2-4} \cline{6-8} 
                                                                                           & \textbf{DeepExpress}            & \textbf{0.1151}   & \textbf{0.0813}  &                                                                                               & \textbf{DeepExpress}            & \textbf{0.1323}   & \textbf{0.0969}  \\ \hline
\multirow{9}{*}{\begin{tabular}[c]{@{}c@{}}Residential/\\ Send\\ Parcels\end{tabular}}     & SARIMA                          & 0.2503            & 0.2158           & \multirow{9}{*}{\begin{tabular}[c]{@{}c@{}}Residential/\\ Receive\\ Parcels\end{tabular}}     & SARIMA                          & 0.2476            & 0.2107           \\ \cline{2-4} \cline{6-8} 
                                                                                           & Prophet                         & 0.2305            & 0.2047           &                                                                                               & Prophet                         & 0.2163            & 0.1808           \\ \cline{2-4} \cline{6-8} 
                                                                                           & SVR                             & 0.2435            & 0.2094           &                                                                                               & SVR                             & 0.2371            & 0.2024           \\ \cline{2-4} \cline{6-8} 
                                                                                           & Xgboost                         & 0.203             & 0.1687           &                                                                                               & Xgboost                         & 0.1959            & 0.1596           \\ \cline{2-4} \cline{6-8} 
                                                                                           & Seq2seq                         & 0.1607            & 0.1256           &                                                                                               & Seq2seq                         & 0.1534            & 0.1237           \\ \cline{2-4} \cline{6-8} 
                                                                                           & Att-Seq2seq                     & 0.1435            & 0.1119           &                                                                                               & Att-Seq2seq                     & 0.1478            & 0.1216           \\ \cline{2-4} \cline{6-8} 
                                                                                           & DA-RNN                          & 0.1428            & 0.1092           &                                                                                               & DA-RNN                          & 0.1489            & 0.1239           \\ \cline{2-4} \cline{6-8} 
                                                                                           & EA-LSTM                         & 0.1406            & 0.1058           &                                                                                               & EA-LSTM                         & 0.1414            & 0.1073           \\ \cline{2-4} \cline{6-8} 
                                                                                           & \textbf{DeepExpress}            & \textbf{0.1239}   & \textbf{0.0894}  &                                                                                               & \textbf{DeepExpress}            & \textbf{0.1247}   & \textbf{0.0919}  \\ \hline
\end{tabular}
\end{table}

In general, a perfect prediction has a value of 0 of RMSE, and a larger value usually indicates a worse prediction. Besides, MAE in Eq. \eqref{eq:MAE} is the average of absolute error which is less sensitive to outliers, and it is given by:

\begin{equation}
\operatorname{MAE}=\frac{\sum_{i=1}^{N}\left|y_{t}-\hat{y}_{t}\right|}{N}
\label{eq:MAE}
\end{equation}

\subsection{Performance Comparison}
\label{PC}

\subsubsection{Overall Performance.}
\label{OP}
To answer \textit{Q1: Does our proposed DeepExpress outperform existing methods in express delivery prediction?}, we compare our model with classical time series prediction models as well as state-of-art deep-learning based models. The experimental results are shown in Table \ref{tab:three}. From the result, we make the following observations. 

Firstly, the deep learning-based models show better performance than that of classical models. This is because the express delivery sequence owns complex non-linear relations and affected by several coupled external features. The deep-learning-based models could easily capture the long-term temporal dependency and fuse multiple features, thus they are more suitable for this prediction problem. 

Secondly, the classic Seq2seq shows the worst performance in the deep learning-based models, because of the lack of attention mechanism limits its performance in the case of long-term sequence dependence as well as multiple features. Att-Seq2seq and DA-RNN show similar performance in most cases, that is because the Att-Seq2seq takes temporal attention to learn temporal dependencies, and DA-RNN uses dual-stage attention to select the most relevant features at each time step. And the EA-LSTM designs a competitive random search the idea of the gradient-based method to solve the attention layer weights, and further improved the performance. However, its performance is not as good as DeepExpress since it weakens the heterogeneous situation in the data. Besides, the impact of features on the sequence are short-term and dynamic changed. Inputting the sequence and features together indiscriminately will cause feature redundancy, which will also affect performance.

Finally, the proposed DeepExpress model outperforms baselines in various application scenarios and achieves the best RMSE and MAE across all types of regions and delivery behaviors. And the results illustrate the complexity of the impact on express delivery. Compared with the classical models, the performance of the DeepExpress has been significantly improved. Because it combines external features with sequence auto-regression, and can simultaneously learn the long-term time dependence and feature impact of the sequence. Compared with the deep learning model, the DeepExpress is still the best, because it considers the heterogeneity of features and can avoid the problem of information loss. Meanwhile, the joint training attention mechanism ensures that the model can learn the complex coupling between sequences and features adaptively.

To conclude, the DeepExpress significantly outperform all baselines and have stronger robustness at the same time.

\subsubsection{Effect of Parameter Setting}
To answer \textit{Q2:What parameters of DeepExpress are important to tune the model?}, we study the effect of parameters $h$,$l$,$k$ with a set of experiments, while keep two parameters fixed and change the other one, and compare the predicted performance.

\begin{figure}[!t]
    \centering
    \subfigure[RMSE-Historical sequence length]{
        \label{his rmse)} 
        \includegraphics[width=0.32\textwidth]{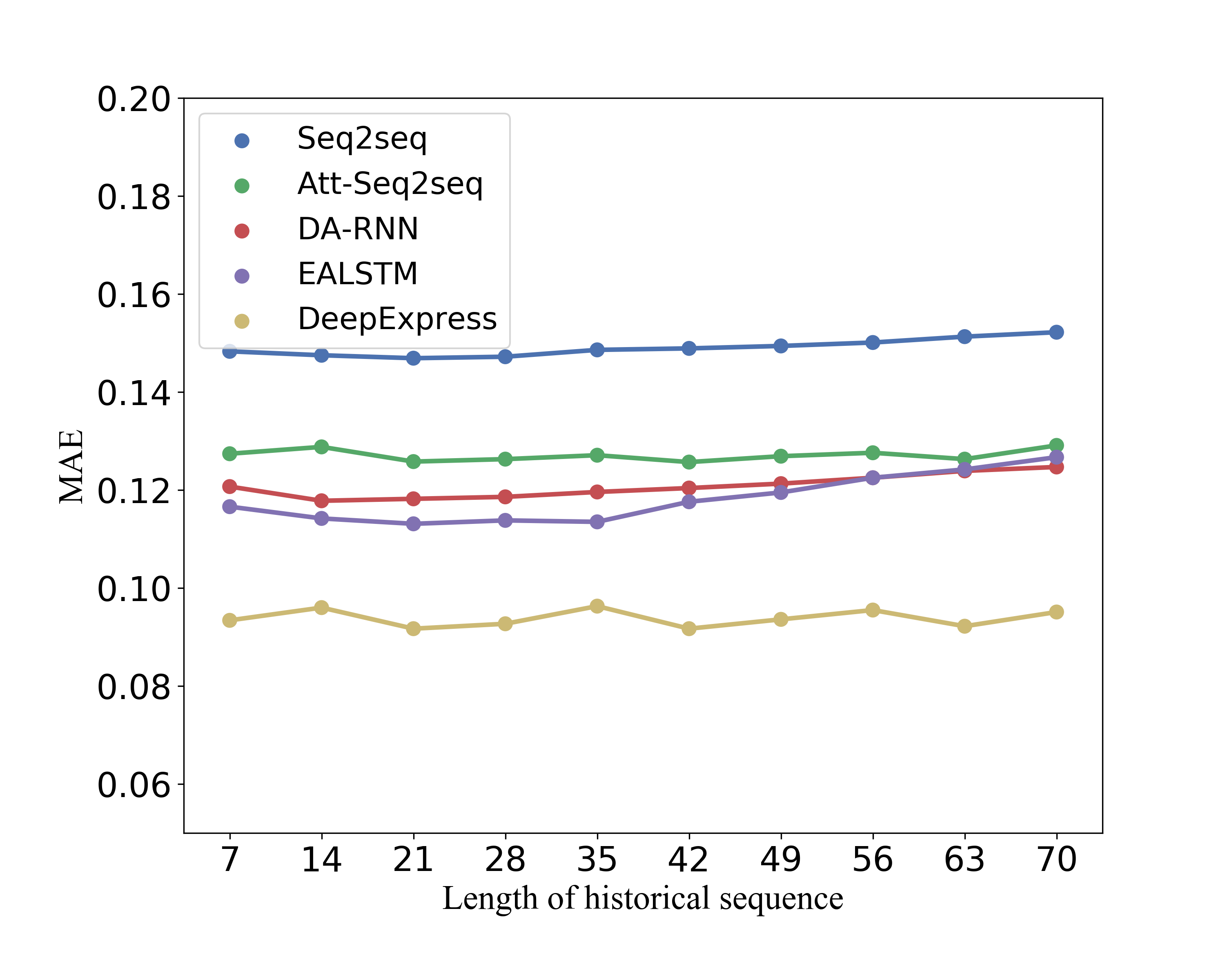}
    }%
    \subfigure[RMSE-Feature time window length]{
        \label{fea rmse)} 
        \includegraphics[width=0.32\textwidth]{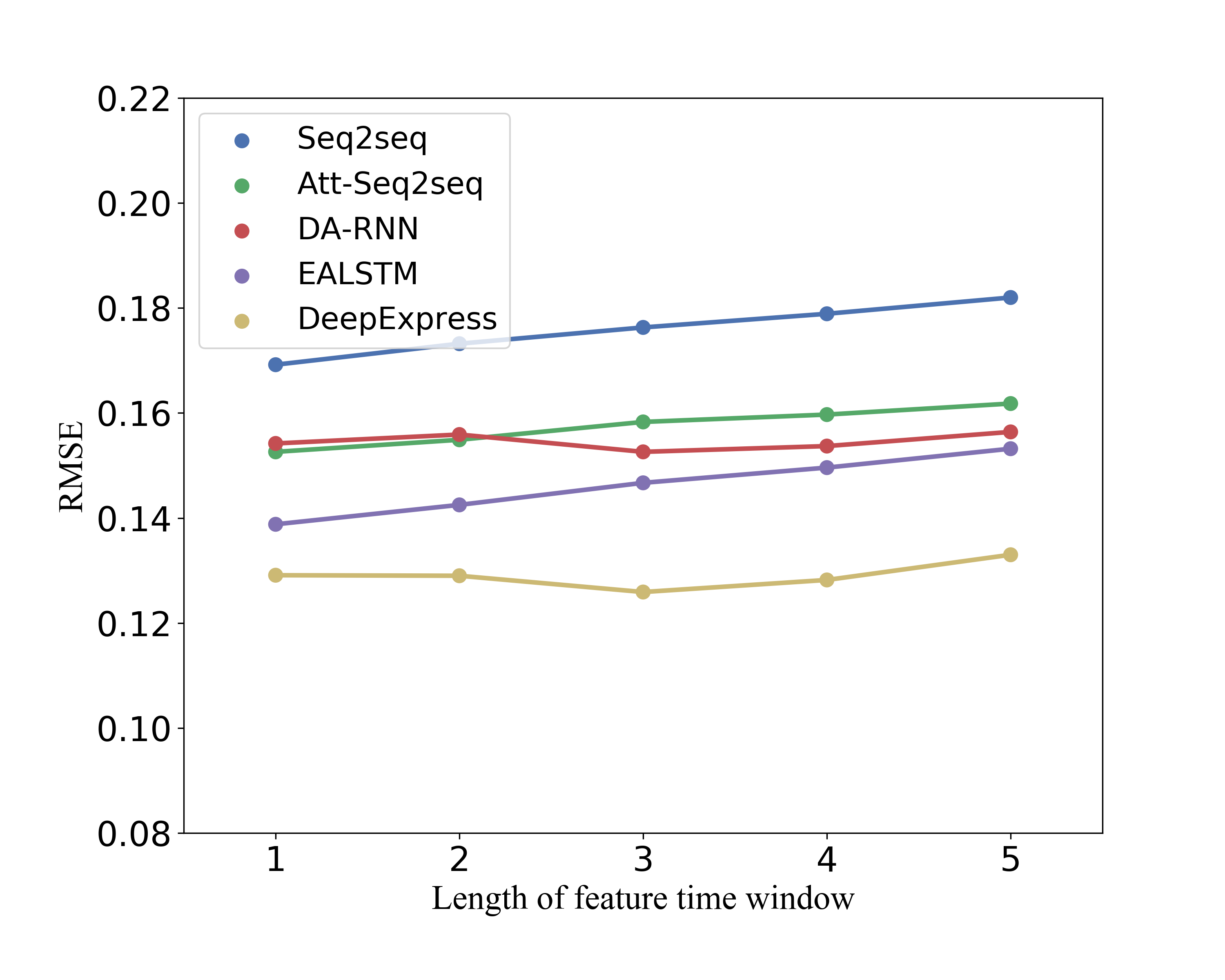}
    }%
    \subfigure[RMSE-Predicted sequence length]{
        \label{pre rmse} 
        \includegraphics[width=0.32\textwidth]{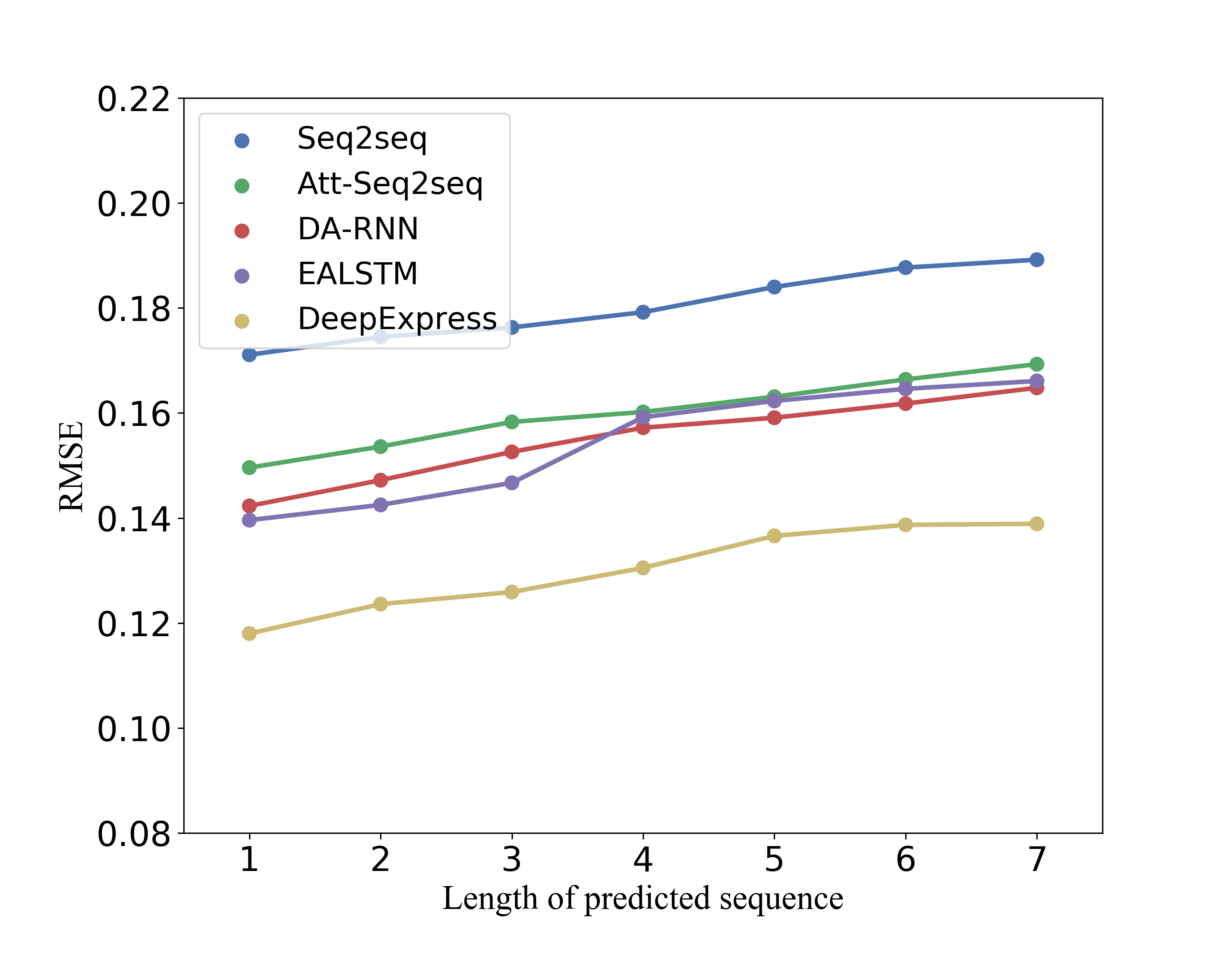}
    }%
    \centering
    
    \centering
    \subfigure[MAE-Historical sequence length]{
        \label{his mae} 
        \includegraphics[width=0.32\textwidth]{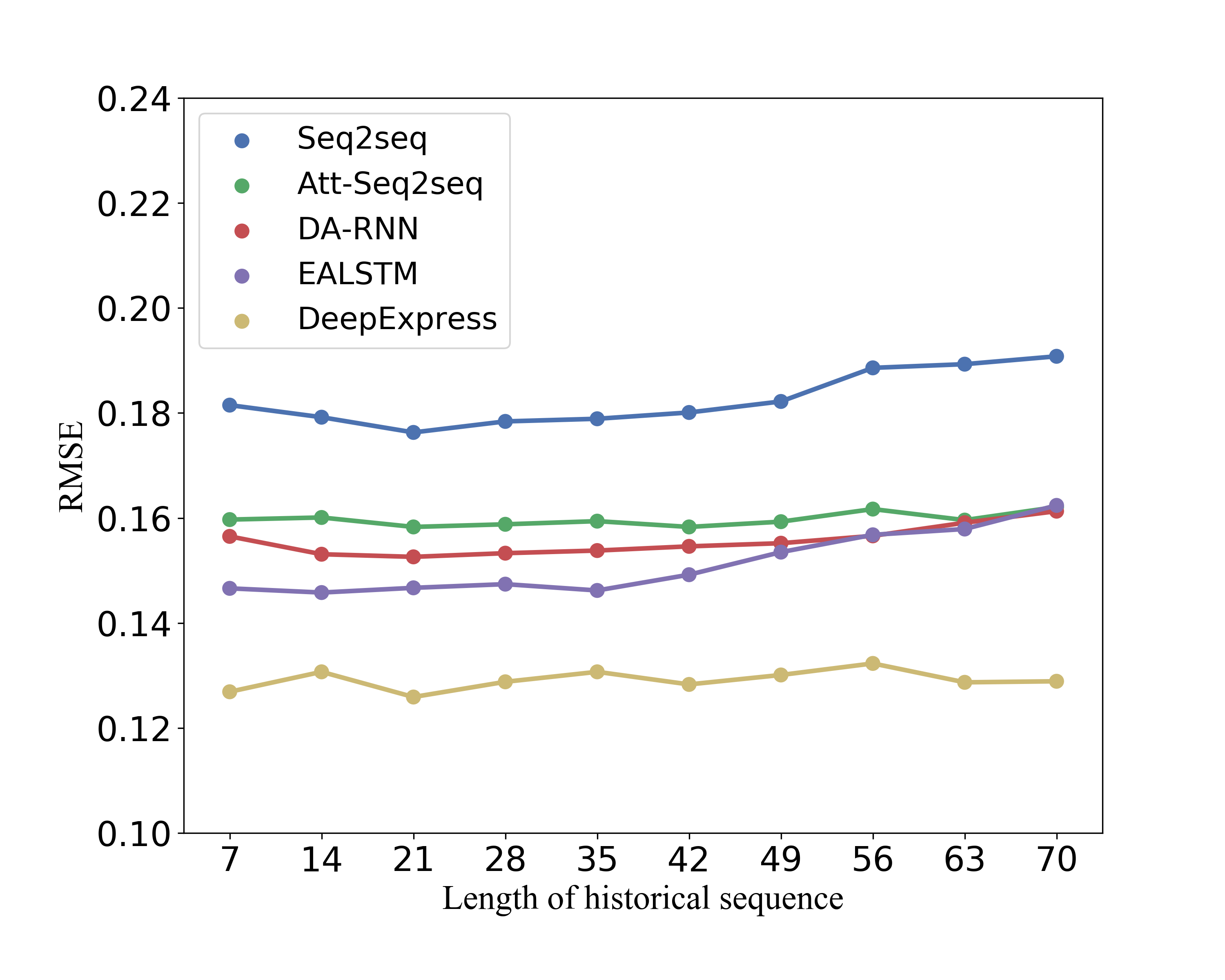}
    }%
    \subfigure[MAE-Feature time window length]{
        \label{fea mae} 
        \includegraphics[width=0.32\textwidth]{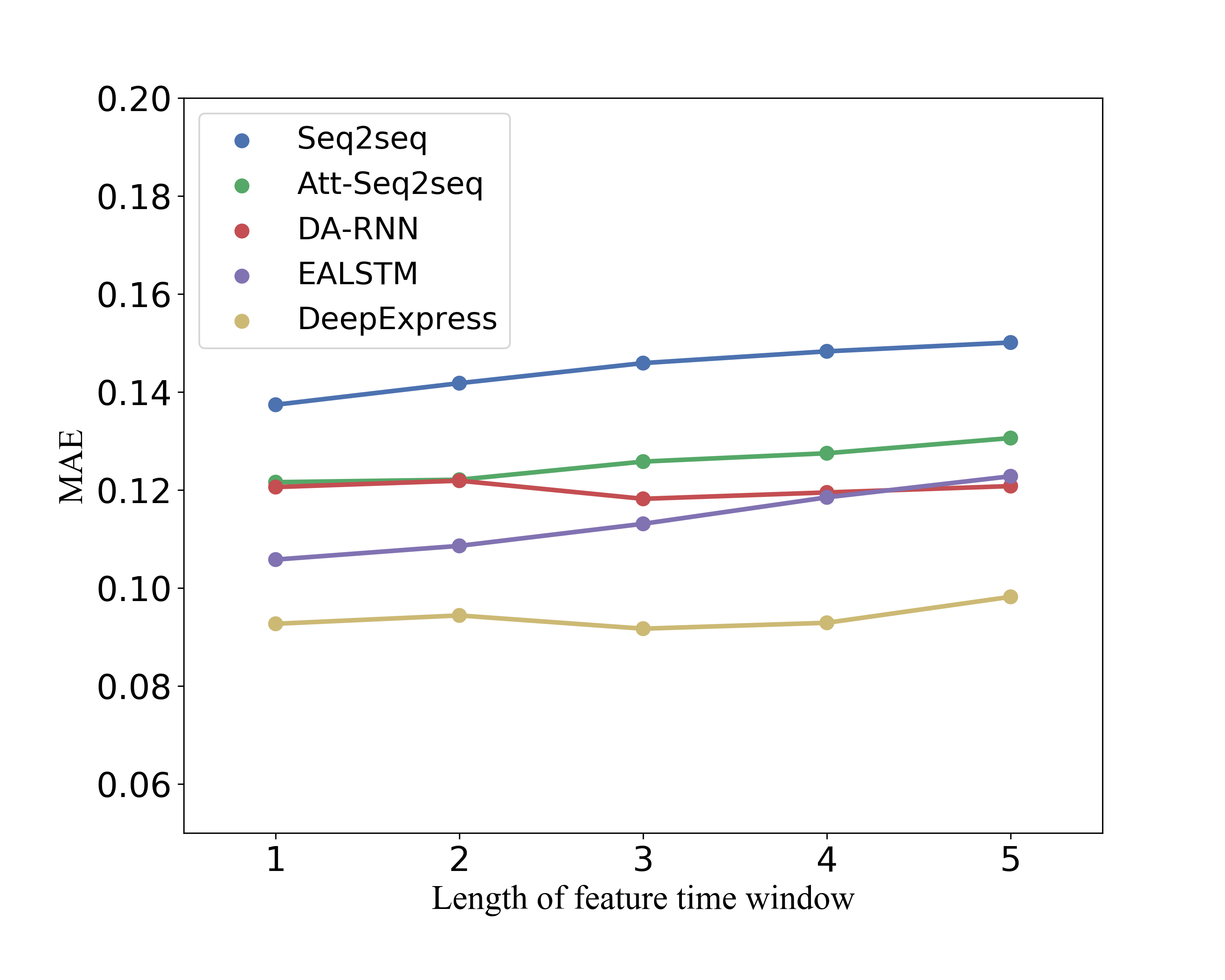}
    }%
    \subfigure[MAE-Predicted sequence length]{
        \label{pre mae} 
        \includegraphics[width=0.32\textwidth]{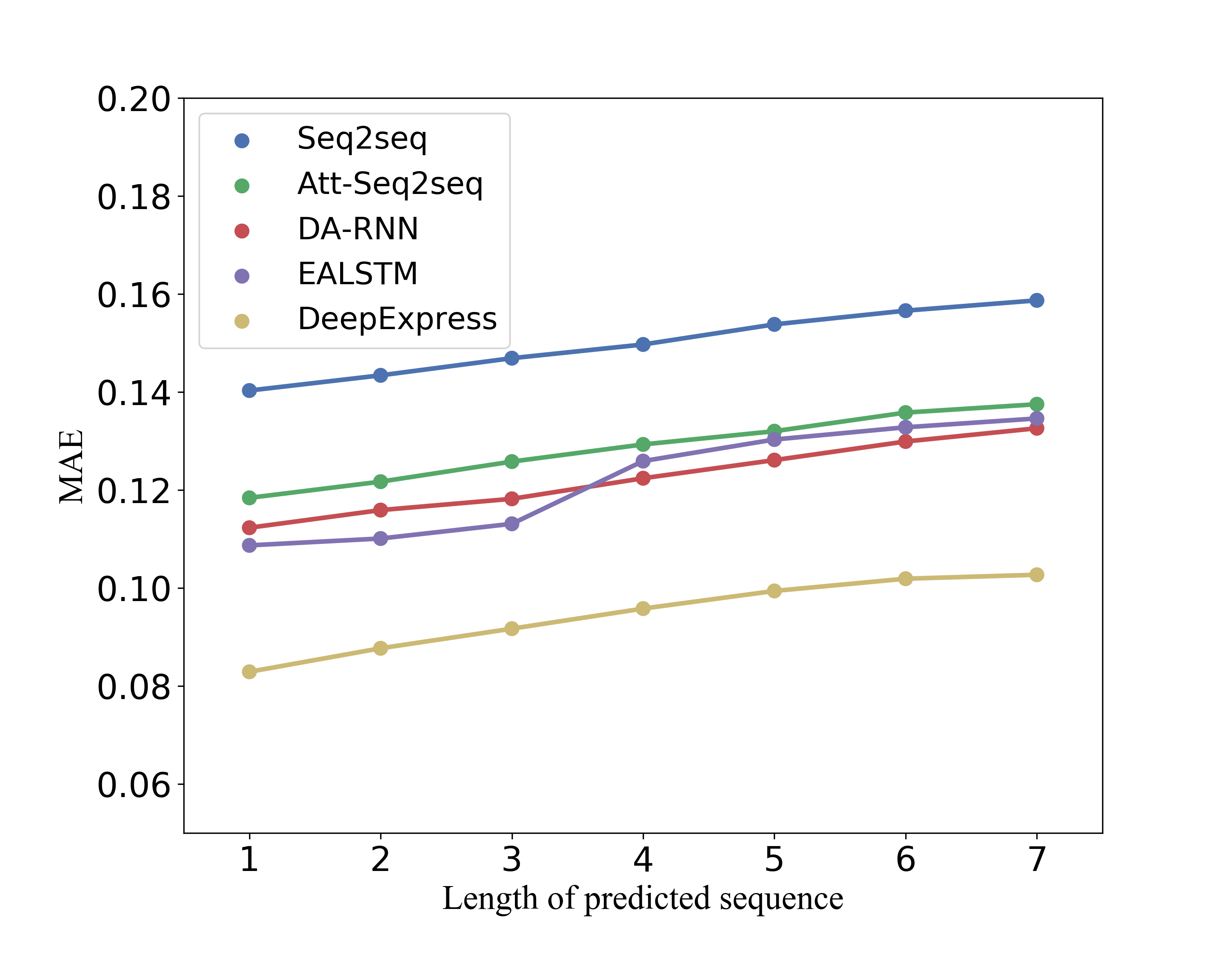}
    }%
    \centering
    
\caption{Performance comparison with all deep learning-based baselines.}
\label{fig:8}
\end{figure}

\textbf{Results on different historical sequence lengths.}
Firstly, we conduct experiments on a fixed predicted sequence length and feature time window length to show the performance comparison of deep learning-based models and the impact of historical sequence length. Specifically, we evaluate the historical sequence length \textit{h} range from 7 to 70 days, and the interval is set as 7 days (i.e., $\mathrm{h} \in\{7,14,21, \ldots, 70\}$). In this way, the length of the historical sequence is an integer multiple of the week which enables the model to obtain complete information for multiple weeks. Considering the overall performance, we set the other parameters as: $l$=3, $k$=3. As shown in Fig. \ref{his rmse)} and Fig. \ref{his mae}, we can observe that the performance has a decreasing trend with all of the approaches except DeepExpress and Att-Seq2seq when the parameter length gets larger. Moreover, the performance of DeepExpress changes periodically as the length increases and achieves the best performance when the length is set as 21, 42 and 63 respectively. It indicates that the temporal attention mechanism of DeepExpress successfully captures the time dependencies so as to keep the model performance. 

\textbf{Results on different feature time window lengths.}
Secondly, we change the length of feature time window, and keep the rest parameters fixed ($h$=21, $k$= 3). According to life experience, the average delivery time usually cost 3 to 5 days, so we set the parameter $l$ varies from 1 to 5 (i.e., $l \in\{1,2,3,4,5\}$). Therefore, the length of feature time window ($2l+ 1$) varies from 3 to 11. Fig. \ref{fea rmse)} and Fig. \ref{fea mae} show the result of the experiments on different lengths of feature time window. We notice that DeepExpress outperforms other models regarding both RMSE and MAE at all lengths of the historical features. And also, it has lower performance degradation when the length increased, which indicates the effectiveness of the feature attention mechanism. Besides, the performance of models first increases and then decreases rapidly, and achieves the best performance when the length equals 3. That is because features like weather and holiday have a hysteresis effect, one day is not enough for a model to learn comprehensive information, while too long may contain a lot of useless information, which affects the predicted results of the model as well.

\textbf{Results on different predicted sequence lengths.}
In comparison to the fixed predicted sequence length, this phase shows experimental results on varying prediction sequence with a fixed input length ($h$=21, $l$=3). Specifically, we limit the length of the predicted sequence to one week. That is because in the real application, the required external feature, weather forecast, will be inaccurate if the sequence is too long, which will further affect model performance. Thus the parameter $k$ is set varies from 1 to 7 days ((i.e., $\mathrm{k} \in\{1,2,3,4,5,6,7\}$).

Fig. \ref{pre rmse} and Fig. \ref{pre mae} show the predicted results. They indicate that the performance of all models decreases as the length of the predicted sequence increases.  Even though, DeepExpress performs better than other approaches in terms of most of the prediction sequence lengths. When the length increases, DeepExpress can still maintain a relatively low RMSE and MAE.

\begin{figure}[!t]
	\centering
	\includegraphics[width=1\linewidth]{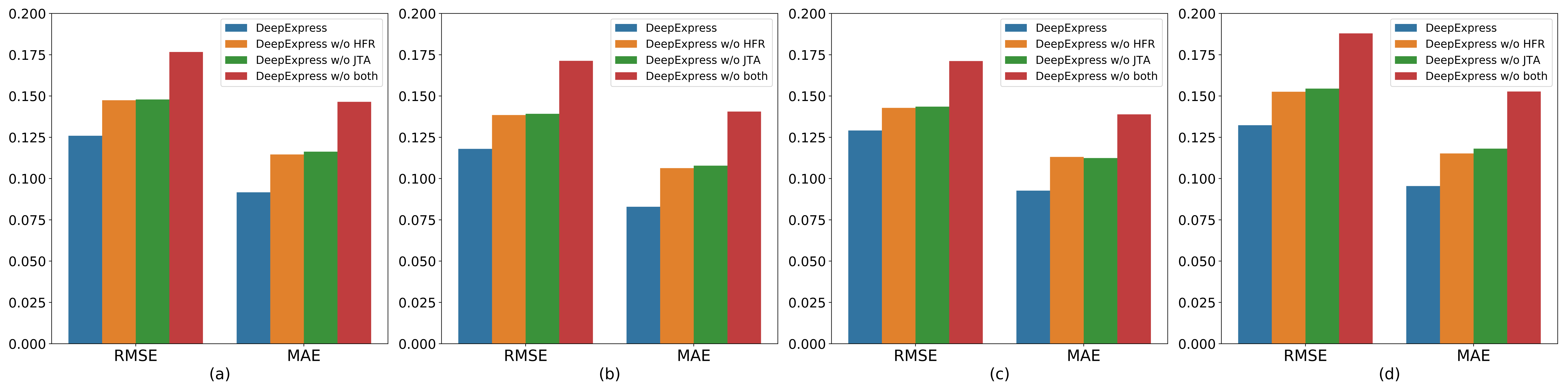}
	\caption{Effects of the HFR and JTA components on DeepExpress, and parameters $h$,$l$,$k$ are set to:$\left(a\right)=\left(21,3,3\right)$, $\left(b\right)=\left(21,1,3\right)$, $\left(c\right)=\left(21,3,1\right)$, and $\left(d\right)=\left(56,3,3\right)$}
	\label{fig:ab}
\end{figure}

\subsubsection{Ablation Study}
To answer \textit{Q3: Weather the proposed joint training attention mechanism successfully captures the complex couplings between the sequence and features?}, we conduct the ablation study by removing the two components manually and comparing the performance with the original DeepExpress.

Specifically, three-fold ablation experiments have been implemented: 1) We remove the Heterogeneous Feature Representation component by treating features equally and directly encoding them into the model (denoted as DeepExpress w/o HFR), which means that we ignore the heterogeneous situation of the data and do not treat features separately. 2) We replace the Joint Training Attention Mechanism by simply averaging the effects of all features and add up the results (denoted as DeepExpress w/o JTA), which means that we do not consider the complex coupling relationships between the sequence and features. 3) Moreover, we remove both the two components simultaneously (denoted as DeepExpress w/o both) to conduct experiments. Experimental results on University dataset are shown in Fig. \ref{fig:ab}.

From the results, we could find out that the performance of DeepExpress decreases when any component is abandoned, because the HFR component takes data heterogeneous situation into consideration and the JTA component captures complex couplings and weigh the relevant features adaptively. DeepExpress w/o HFR lost feature heterogeneous information and DeepExpress w/o JTA fail to capture the dynamic change effects of the features, so they both obtain worse performance. DeepExpress w/o both shows the worst results as it does not consider either the heterogeneous features or complex couplings in the prediction process. 

\section{Conclusion}
\label{sec:CON}
In this paper, we propose a data-driven approach to predict the express delivery sequence. Based on the domain knowledge about express delivery behaviors, we design a novel express delivery sequence prediction model, namely DeepExpress, to learning complex coupling between sequence and features. It consists of three novel characteristics unavailable in previous methods: 1) an express delivery seq2seq learning, which combines sequence auto-regressive process with external features' effect for time series modeling; 2) a carefully designed heterogeneous feature representation, which adaptively maps heterogeneous features to a unified hidden representation while preserving the original information; and 3) a novel joint training attention mechanism that simultaneously learns sequence temporal dependencies and coupled feature effects from dynamic changes, and assigns weights adaptively. Benefited from heterogeneous representation and joint training of coupled features and sequence, DeepExpress outperforms the existing express delivery sequence prediction methods. Experiments on real-world express delivery data demonstrate the effectiveness of DeepExpress in terms of prediction accuracy.

The present DeepExpress is designed for express delivery prediction problems, yet it is also suitable for general time series prediction and our future work is to employ DeepExpress for public datasets to verify its versatility. Moreover, we will extend our method to solve the problem of multi-sequence couplings through multi-variate time series.

%
\begin{acks}
This work was partially supported by the National Key R\&D Program of China (2019YFB1703901) and  the National Natural Science Foundation of China (No. 61772428,61725205,61902320,61972319).
\end{acks}

%

\bibliographystyle{unsrt}
\bibliography{bibliography}

%
\appendix

\end{document}